\documentclass[lettersize,journal]{IEEEtran}
\usepackage{amsmath,amsfonts}
\usepackage{algorithmic}
\usepackage{algorithm}
\usepackage{array}
\usepackage[caption=false,font=normalsize,labelfont=sf,textfont=sf]{subfig}
\usepackage{textcomp}
\usepackage{stfloats}
\usepackage{url}
\usepackage{tabularx}
\usepackage{tabularray}
\usepackage[dvipsnames]{xcolor}
\usepackage{verbatim}
\usepackage{graphicx}
\usepackage{cite}
\usepackage{hyperref}
\usepackage{bbding}
\usepackage{pifont}
\usepackage{wasysym}
\usepackage{amssymb}
\UseTblrLibrary{booktabs, caption}
\usepackage{tikz}
\usepackage{xcolor}
\usepackage{booktabs}% http://ctan.org/pkg/booktabs
\newcommand{\tabitem}{{\textbullet}\hspace{0.5em}}
\usepackage{enumitem}
\usepackage{wasysym}
\newcommand{\graycircle}{\tikz\draw[gray,fill=gray] (0,0) circle (.75ex);}
\newcommand{\navybluecircle}{\tikz\draw[NavyBlue,fill=NavyBlue] (0,0) circle (.75ex);}

\newcommand{\apricotcircle}{\tikz\draw[Apricot,fill=Apricot] (0,0) circle (.75ex);}

\newcommand{\yellowgreencircle}{\tikz\draw[YellowGreen,fill=YellowGreen] (0,0) circle (.75ex);}
\newcommand{\emptycircle}{\tikz\draw[black] (0,0) circle (.75ex);}

\newcommand{\halfblackcircletwo}{\tikz{
    \fill[black] (0,0) -- (90:.75ex) arc[start angle=90, end angle=270, radius=.75ex] -- cycle; % Fill left half with black
    \draw (0,0) circle (.75ex); % Draw the outline of the circle
}}

\begin{document}
%\title{Large Language Model (LLM) Meets AI Chatbot: A Comprehensive Survey on ChatGPT and Beyond}
%\title{A Comprehensive Survey on Chatbot Powered by Large Language Model}
%\title{A Comprehensive Survey on AI Chatbots Powered by Large Language Models}
%\title{Large Language Model (LLM) Meets AI Chatbot: A Comprehensive Survey on Conversational Agents}
\title{A Complete Survey on LLM-based AI Chatbots}
\author{Sumit Kumar Dam, Choong Seon Hong,~\IEEEmembership{Fellow,~IEEE}, Yu Qiao,~\IEEEmembership{Student Member,~IEEE,}\\and Chaoning Zhang,~\IEEEmembership{Senior Member,~IEEE}
\thanks{Sumit Kumar Dam, Choong Seon Hong, Yu Qiao, and Chaoning Zhang are with the School of Computing, Kyung Hee University, Yongin-si 17104, Republic of Korea (email: skd160205@khu.ac.kr; cshong@khu.ac.kr; qiaoyu@khu.ac.kr; chaoningzhang1990@gmail.com). \textit{(Sumit Kumar Dam and Choong Seon Hong contributed equally to this work.)} \textit{(Corresponding author: Chaoning Zhang.)}}
%\thanks{$^{\dagger}$Equal Contribution}
%\thanks{$^{*}$Corresponding Author}
%\vspace{+1mm}
}
        % <-this % stops a space
% \thanks{This paper was produced by the IEEE Publication Technology Group. They are in Piscataway, NJ.}
% <-this % stops a space
% \markboth{Journal of \LaTeX\ Class Files,~Vol.~14, No.~8, August~2021}
\maketitle
\begin{abstract}
The past few decades have witnessed an upsurge in data, forming the foundation for data-hungry, learning-based AI technology. Conversational agents, often referred to as AI chatbots, rely heavily on such data to train large language models (LLMs) and generate new content (knowledge) in response to user prompts. With the advent of OpenAI’s ChatGPT, LLM-based chatbots have set new standards in the AI community. This paper presents a complete survey of the evolution and deployment of LLM-based chatbots in various sectors. We first summarize the development of foundational chatbots, followed by the evolution of LLMs, and then provide an overview of LLM-based chatbots currently in use and those in the development phase. Recognizing AI chatbots as tools for generating new knowledge, we explore their diverse applications across various industries. We then discuss the open challenges, considering how the data used to train the LLMs and the misuse of the generated knowledge can cause several issues. Finally, we explore the future outlook to augment their efficiency and reliability in numerous applications. By addressing key milestones and the present-day context of LLM-based chatbots, our survey invites readers to delve deeper into this realm, reflecting on how their next generation will reshape conversational AI.
\end{abstract}
\vspace{+2mm}
\section{Introduction}
\begin{figure}[htbp]
\centerline{\includegraphics[width=\linewidth]{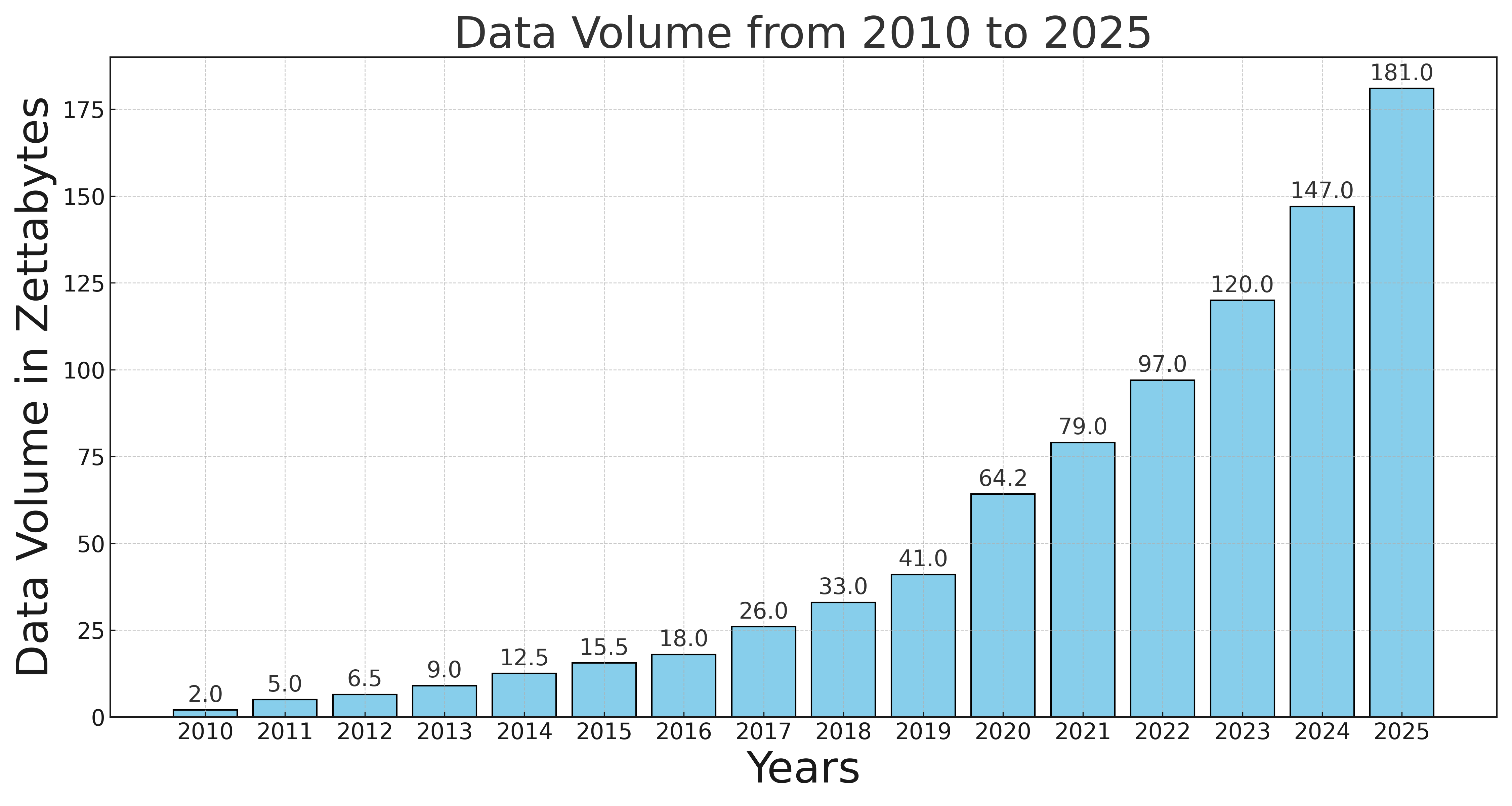}}
\caption{Increase in data volume over the years \cite{WorldDataVolume}.}
\label{fig:Figure0}
%\vspace{-2mm}
\end{figure}

\IEEEPARstart{T}{HE} exponential growth of data in recent years has transformed the world of digital information. In 2023, the total amount of data created, captured, copied, and consumed globally reached around 120 zettabytes, and it will probably go up to 147 zettabytes by 2024, with expectations to surpass 180 zettabytes by 2025 \cite{WorldDataVolume}. Fig. \ref{fig:Figure0} illustrates the%
increase in data volume from 2010 to 2023, with projected values for 2024 and 2025. This rapid expansion of the data ecosystem has paved the way for groundbreaking innovations in artificial intelligence (AI), leading to the development of several machine learning models. Among these, large language models (LLMs) have emerged as a prominent subset \cite{zhao2023survey} thanks to their exceptional ability to understand, generate, and manipulate human language \cite{brown2020language}.\\
\begin{comment}
In the past few decades, the field of artificial intelligence (AI) has advanced significantly, leading to the development of several machine learning models. Among these, large language models (LLMs) have emerged as a prominent subset \cite{zhao2023survey} thanks to their exceptional ability to understand, generate, and manipulate human language \cite{brown2020language}.\\  
\end{comment}
\begin{figure}[t]
\centerline{\includegraphics[width=\linewidth]{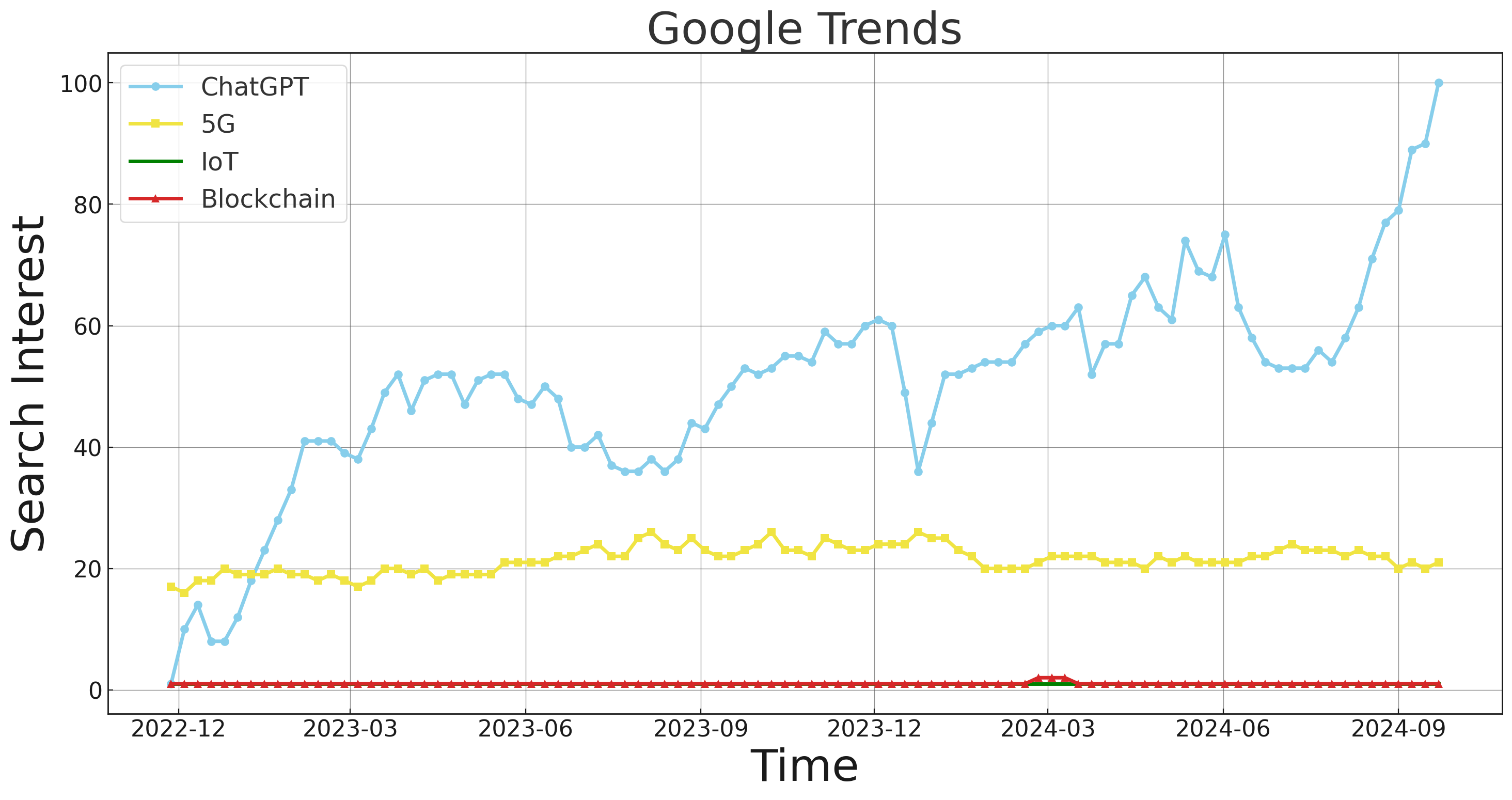}}
\caption{Google search interest over time \cite{GoogleTrends2024IoT}.}
\label{fig:Figure1}
\vspace{-6mm}
\end{figure}
\begin{table*}[!ht]
\centering
\captionsetup{labelsep=newline,justification=centering}
\caption{Summary of Chatbot Literature}
\label{tab:TableNew}
\begin{tblr}{
  width = \textwidth,
  colspec = {|Q[c,m,0.0663\linewidth]|Q[c,m,0.05427\textwidth]|Q[c,m,0.05427\linewidth]|Q[c,m,0.1064\linewidth]|Q[c,m,0.1064\linewidth]|Q[c,m,0.0555\linewidth]|Q[c,m,0.0555\linewidth]|Q[c,m,0.0555\linewidth]|Q[j,m,0.2272\linewidth]|},
  row{1} = {font=\bfseries, c},
  row{1-16} = {font=\linespread{1.1}\selectfont}, % Adjust line spread for all cells, if needed
  hlines,
  cell{1}{2} = {c=2}{},
  cell{2}{2} = {c=1}{halign=c,valign=m},
  cell{2}{3} = {c=1}{halign=c,valign=m},
  cell{1}{4} = {c=2}{},
  cell{2}{4} = {c=1}{halign=c,valign=m},
  cell{2}{5} = {c=1}{halign=c,valign=m},
  cell{1}{6} = {c=3}{},
  cell{2}{6} = {c=1}{halign=c,valign=m},
  cell{2}{7} = {c=1}{halign=c,valign=m},
  cell{2}{8} = {c=1}{halign=c,valign=m},
  cell{1}{1} = {r=2}{halign=c,valign=m},
  cell{1}{9} = {r=2}{halign=c,valign=m},
}
Referenced Article         & \SetCell[c=2]{c} Chatbot(s) Covered        && \SetCell[c=2]{c} Application(s) Scope          && \SetCell[c=3]{c} Challenge(s) Discussed &&& Remark(s) \\
                           &  Single & Multiple        &  Monodisciplinary   & Multidisciplinary & Tehcnical &  Ethical & Misuse & \\
\cite{kooli2023chatbots}  &  \graycircle  & \textbf{-}    & \textbf{-}  & \yellowgreencircle  & \emptycircle & \CheckmarkBold & \CheckmarkBold & {\tabitem Limits discussion to ChatGPT, with no mention of other chatbots such as BARD, Bing Chat, etc.\\
\tabitem Overlooks the technical issues.}\\

\cite{tlili2023if}  &  \graycircle  & \textbf{-}    & \graycircle & \textbf{-}  & \emptycircle & \CheckmarkBold & \CheckmarkBold & {\tabitem Does not address technical issues.\\
\tabitem Explores ethical concerns and misuse cases under user experience, resulting in underdeveloped content.}\\

\cite{koubaa2023exploring}  &  \graycircle  & \textbf{-}    & \textbf{-}  & \navybluecircle  & \halfblackcircletwo & \CheckmarkBold & \halfblackcircletwo & \tabitem Inadequately addresses the technical issues and misuse cases, lacking depth and detail in the discussion.\\

\cite{sallam2023chatgpt}  &  \graycircle  & \textbf{-} & \graycircle  & \textbf{-}  & \CheckmarkBold & \CheckmarkBold & \CheckmarkBold & \tabitem Lacks structured categorization of key issues, causing difficulties for readers to find specific information.\\ %The article discusses all the limitations but they are not well-categorized which makes it difficult for the readers having interest in a specific segment.

\cite{lo2023impact}  &  \graycircle  & \textbf{-} & \graycircle  & \textbf{-}  & \halfblackcircletwo & \halfblackcircletwo & \halfblackcircletwo & {\tabitem Only considers the initial release of ChatGPT (v3.5).\\
\tabitem Lacks categorization and depth in discussing educational issues.}\\ %First, the study only considers the initial version of ChatGPT (original release of ChatGPT). There is no proper categorization of the issues associated with ChatGPT in the education domain. Besides, they are also not discussed in full potential.\\

\cite{hosseini2023exploratory}  &  \graycircle  & \textbf{-} & \textbf{-}  & \apricotcircle & \halfblackcircletwo & \halfblackcircletwo & \halfblackcircletwo & \tabitem Lacks a taxonomy for applications and challenges across sectors.\\ %& The uses in different sectors are not well-categorized. No good taxonomy. Also, There is no proper categorization of the issues associated with ChatGPT in different domains. The authors discuss on different issues like hallucination, transparency, bias, but a lot of essential discussions are missing.\\

\cite{ray2023chatgpt}  &  \graycircle  & \textbf{-} & \textbf{-}  & \navybluecircle & \CheckmarkBold & \CheckmarkBold & \CheckmarkBold & {\tabitem Relies heavily on narrative content that lacks analytical depth.\\
\tabitem Lacks sufficient visual aids (e.g., figures, graphs, bar charts, etc.), complicating data interpretation and reducing reader engagement.} \\ %& The article may be using a lot of narrative or descriptive content, which while informative, might lack analytical depth or critical evaluation. This can make the content feel less actionable or insightful, particularly for readers looking for specific guidance or decision-making information. The lack of tables, graphs, or other visual aids can make it harder for readers to quickly interpret or analyze the data and trends that are being discussed. Visuals are not only engaging but also serve as a quick reference point for understanding complex information.\\

\cite{dwivedi2023so}  &  \graycircle  & \textbf{-} & \textbf{-}  & \navybluecircle & \CheckmarkBold & \CheckmarkBold & \CheckmarkBold & \tabitem Contains substantial overlapping insights from 43 contributions by experts, leading to an unorganized and excessively lengthy document.\\  %& The article being an opinion paper has too many redundant information as the paper is a collection of 43 contributions from experts from different fields. On so many occassions, their opinions overlapped which creates redundant, duplicate, and unnecessary lengthy paper. The paper doesn't doesn't do anything to organize them. There is no categorization.\\

\cite{sohail2023decoding}  &  \graycircle  & \textbf{-} & \textbf{-}  & \navybluecircle & \CheckmarkBold & \CheckmarkBold & \CheckmarkBold & \tabitem Lacks fine-grained categorization of applications and challenges.\\  %& Although the article has touched upon all the possible sectors, unlike the Future possibilities, the applications and challenges lack fine-grained categorization.\\

\cite{santos2023enhancing}  &  \textbf{-}  & \yellowgreencircle & \graycircle  & \textbf{-} & \emptycircle & \emptycircle & \emptycircle & \tabitem Lacks discussion on key issues. \\  %& The article doesn't shed light on any of the issues of both chatbots, whether there was any hallucination. It is important since it is doing the case study on Chemistry. Whether the chatbots generate any type of wrong information or not, is not reported.\\

Our Survey  &  \textbf{-}  & \navybluecircle & \textbf{-} & \navybluecircle & \CheckmarkBold & \CheckmarkBold & \CheckmarkBold & {\tabitem Covers a wide range of chatbots beyond ChatGPT, including BARD, Bing Chat, Claude, and others.\\
\tabitem Provides a detailed taxonomy of applications and challenges, each divided into distinct sub-categories.
} \\  %& The article doesn't shed light on any of the issues of both chatbots, whether there was any hallucination. It is important since it is doing the case study on Chemistry. Whether the chatbots generate any type of wrong information or not, is not reported.\\

%From  &  \tikz\draw[darkgray,fill=darkgray] (0,0) circle (.75ex);  & Challenge    & \textbf{-}  & \CheckmarkBold   \\
%From an   & \tikz\draw[NavyBlue,fill=NavyBlue] (0,0) circle (.8ex);          & Lack    & \textbf{-}  & \CheckmarkBold   \\
%From     & \tikz\draw[SkyBlue,fill=SkyBlue] (0,0) circle (.8ex);       & Challenge       & \CheckmarkBold & \textbf{-}    \\
%From     & \tikz\draw[YellowGreen,fill=YellowGreen] (0,0) circle (.8ex);       & Challenge       & \CheckmarkBold & \textbf{-}    \\
\end{tblr}
\begin{description}
\item *~\textit{\CheckmarkBold (Fully discussed); \halfblackcircletwo~(Partially discussed); \emptycircle~(Not discussed); \textbf{-} (Not applicable);}
\item *~\textit{\graycircle~(Single chatbot or monodisciplinary); \yellowgreencircle~(Two chatbots or two disciplines);}
\item *~\textit{\apricotcircle~(Three chatbots or three disciplines); \navybluecircle~(More than three chatbots or more than three disciplines).}
\end{description}
\vspace{-6.35mm}
\end{table*}
\indent In the era of AI-powered chatbots \cite{lin2023review, khosravi2023chatbots, alazzam2023artificial}, LLMs have become instrumental in powering their conversational capabilities and facilitating human-like interactions \cite{zhao2023survey, koubaa2023exploring}. The substantial rise in data and advancements in computational knowledge have improved the functionality of LLM-based chatbots, making them increasingly popular and widely adopted across various sectors. Their ability to understand and respond in human language with never-before-seen context relevance and accuracy, coupled with the capacity to handle vast information streams, makes them essential tools in areas like education \cite{baidoo2023education, song2022impact, lee2022developing}, research \cite{aydin2023google, macdonald2023can, girotra2023ideas}, healthcare \cite{ayanouz2020smart, athota2020chatbot, sallam2023chatgpt}, and many others \cite{belzner2023large, lakkaraju2023can, ChatbotOtherUses}. Given the vast potential and promising possibilities of LLM-based chatbots, their growing usage and required optimization pose several challenges that call for thorough research and evaluation. This need becomes more apparent as the domain of LLM-based chatbots rapidly expands, leading to an overwhelming amount of research literature for scholars, professionals, and newcomers alike. Therefore, our work provides a timely and complete survey of LLM-based chatbots in response to these evolving needs.\\
\indent Before the emergence of LLMs and LLM-based chatbots, conversational AI faced several challenges. Early chatbots had limited contextual understanding and domain specificity. They often provided inaccurate responses. Lack of sophisticated language comprehension limited their ability to interact in a human-like manner, which led to robotic and disjointed user experiences. Scalability across various industries was also problematic, as handling vast information streams with real-time responsiveness proved challenging. The advent of LLMs revolutionized chatbots and started a new era of AI-driven interactions. In March 2023, OpenAI unveiled its latest marvel, GPT-4 (also known as ChatGPT Plus \cite{openai2023gpt4}), following the buzz generated since the debut of ChatGPT 3.5 in November 2022 \cite{bahrini2023chatgpt, zhang2023one}. Fig. \ref{fig:Figure1} illustrates the exponential rise in popularity of ChatGPT (in blue) since its initial release \cite{GoogleTrends2024IoT}, showcasing its dominance over other widespread technologies such as 5G (in yellow), IoT (in green), and Blockchain (in red). Its innovative capabilities have been met with an unprecedented surge in popularity, highlighting a new chapter in AI-driven communication. In a related development, Google announced BARD \cite{GoogleBARD}, its first LLM-based chatbot, on February 6, followed by early access on March 21 \cite{GoogleBARDWiki}. Additionally, there are numerous other LLM-based chatbots in the works. Acknowledging the profound impact of these technologies, this survey aims to provide a distilled, up-to-date overview of LLM-based chatbots, including their development, industry-wide applications, key challenges, and strategies to improve their effectiveness and reliability. Our goal is to integrate these diverse studies into a well-organized survey that will facilitate an in-depth understanding of LLM-based chatbots and give readers a guide for their future research.
\vspace{-4.6mm}
\subsection{Existing Surveys, Reviews, and Case Studies}
Several articles have reviewed the wide-ranging applications of LLM-based chatbots, highlighting their significant impact and the complex challenges they pose across various sectors. Here, we discuss some of these articles and demonstrate how our survey extends and differs from them.\\
\indent \cite{kooli2023chatbots} explores the use of AI and chatbots in the academic field and their influence on research and education from an ethical perspective. It investigates the impact of these technologies on educational assessment integrity and their potential to transform academic research. In addition, it recommends effective solutions to alleviate the ethical challenges and possible misuse of these tools in the education and research domains. \cite{tlili2023if} conducts a case study on how ChatGPT elevates online learning. The findings suggest that students favor these agents for their educational activities, citing a more interactive and engaging learning environment. Koubaa et al. \cite{koubaa2023exploring} provide a detailed review of ChatGPT’s technical novelties. Following this, they develop a unique taxonomy in their survey for research categorization and explore ChatGPT’s applications across various fields. Additionally, they highlight significant challenges and avenues for future exploration. \cite{sallam2023chatgpt} offers a systematic review of ChatGPT in healthcare, focusing on education, research, and practice. The author outlines ChatGPT’s potential to revolutionize scientific writing and personalized learning. The review critically analyzes the benefits while acknowledging significant concerns, such as ethical and accuracy issues. Another review article \cite{lo2023impact} evaluates ChatGPT’s impact on education, noting its varied performance across subjects such as Economics, Programming, Law, Medical Education, and Mathematics, among others. The paper highlights the tool’s potential and challenges, like accuracy concerns and plagiarism, suggesting updates to assessment methods and educational policies for responsible use. In \cite{hosseini2023exploratory}, the authors conduct an exploratory survey utilizing virtual and in-person feedback to analyze the impact of ChatGPT in education, healthcare, and research. The survey demonstrates how ChatGPT can improve personalized learning, clinical tasks, and research efficiency. They also address major ethical and practical concerns, suggesting careful AI deployment with robust ethical guidelines to navigate these challenges. In a similar context, \cite{ray2023chatgpt} provides a comprehensive analysis of ChatGPT, focusing on its evolution, diverse applications, and key challenges. Unlike \cite{hosseini2023exploratory}, which employs surveys for direct feedback, \cite{ray2023chatgpt} aggregates findings from existing studies to assess the influence and challenges of ChatGPT, offering a more generalized perspective without engaging in primary data collection. Exploring further, \cite{dwivedi2023so} and \cite{sohail2023decoding} delve into ChatGPT’s broader multidisciplinary applications. \cite{dwivedi2023so} gathers insights across multiple disciplines to assess its effects in sectors ranging from marketing to education and healthcare, whereas \cite{sohail2023decoding} introduces a taxonomy of ChatGPT research, detailing its applications in domains like healthcare, finance, and environmental science. Besides, both papers address fundamental challenges regarding ethical considerations and practical deployment. Another recent article \cite{santos2023enhancing} uses a single-case study approach to evaluate the effectiveness of ChatGPT and Bing Chat in Chemistry education. The research analyzes extensive interactions between these tools and a simulated student to improve creativity, problem-solving, and personalized learning. The findings show that both chatbots act as valuable ‘agents-to-think-with.’ However, ChatGPT notably outperforms Bing Chat in delivering more comprehensive and contextually accurate responses across various scientific topics.\\
\indent Different from existing works, our survey expands beyond the typical focus on specific chatbots like ChatGPT, covering a wide range of models, including BARD, Bing Chat, and Claude. Moreover, we explore applications across multiple domains and discuss various challenges, each detailed in several sub-categories. Table \ref{tab:TableNew} summarizes the findings of the discussed articles, facilitating a comparative understanding of their contributions.\\
%%%%%%%%%%%%%%%%%%%%%%%%%%%%%%%
\begin{figure*}[htbp]
\centerline{\includegraphics[width=0.98\linewidth]{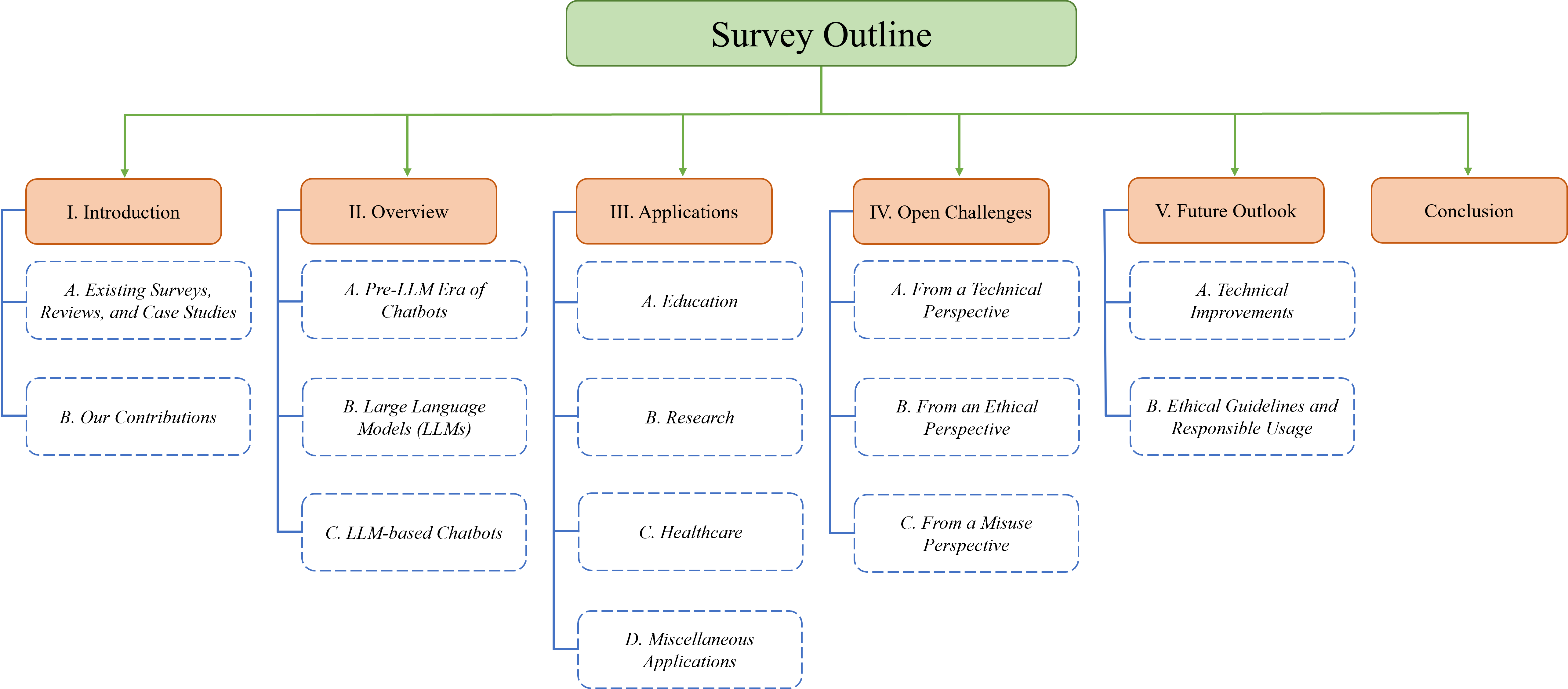}}
\caption{Survey outline.}
\label{fig:Picture4Alt9}
\vspace{-4mm}
\end{figure*}
\vspace{-7mm}
\subsection{Our Contributions}
Our survey aims to answer the following questions:
\begin{itemize}
  \item How have chatbots evolved from simple automated systems to the LLM-based variants we see today, and what foundational advancements in LLMs have redefined chatbot capabilities since the pre-LLM era?
  \item What are the key applications of LLM-based chatbots across different sectors, and how do they impact the operational dynamics and user interactions within these domains?
  \item What challenges emerge with the widespread use of LLM-based chatbots, and how do they affect their performance and reliability?
  \item What technical improvements are essential for LLM-based chatbots, and how will the implementation of ethical guidelines ensure their responsible usage?
\end{itemize}

\indent In addressing these questions, we offer a comprehensive overview of the history of chatbots. Additionally, we discuss the foundations of LLMs, highlighting the transformer-based self-attention mechanisms and the innovative features in GPT models, such as in-context learning and chain-of-thought (CoT) prompting. Then, we provide a detailed taxonomy of LLM-based chatbots, organizing them by their functionalities and applications in sectors like education, research, and healthcare. We also acknowledge their growing significance in software engineering and finance. Next, we look into the open challenges from technical aspects, covering issues from knowledge recency to hallucination, alongside ethical considerations like data transparency, bias, privacy risks, and unfairness. We then conclude with misuse perspectives, highlighting concerns regarding academic misuse, over-reliance, and the distribution of wrong information. Finally, we discuss the future outlook for LLM-based chatbots, from technical improvements like model optimization to complying with ethical guidelines and promoting responsible use across various domains. Our contributions are summarized as follows:
\begin{itemize}
    \item Unlike most articles that concentrate on specific chatbots or their limited aspects, our survey covers a variety of LLM-based models, including ChatGPT, BARD, Bing Chat, and many others.
    \item While the majority of articles focus on a single chatbot applied to one or multiple domains without detailed categorization, our survey extends to a wide array of chatbots across various application domains. We provide a detailed taxonomy of applications, offering a structured and in-depth exploration of how different chatbots perform across sectors like education, research, healthcare, software engineering, and finance.
    \item We discuss several open challenges from technical, ethical, and misuse perspectives. In addition, we frame our discussion around knowledge and data, the two central pillars of LLMs. This approach illustrates the dynamic interplay between chatbots' interaction with extensive training data and their subsequent generation of new content (knowledge).
\end{itemize}

The rest of the survey is organized as follows: Section \ref{sec:Overview} covers the foundational years of chatbots, the rise of LLMs, and an overview of LLM-based chatbots. Section \ref{sec:Applications} highlights the applications of these chatbots in education, research, and healthcare. It also covers miscellaneous applications, such as software engineering and finance. Section \ref{sec:Challenges} delves into the inherent challenges of these chatbots, while Section \ref{sec:Outlook} explores the future outlook in this field. Finally, Section \ref{sec:Conclusion} concludes the survey, summarizing its key findings and overall contributions. The outline of our survey is shown in Fig. \ref{fig:Picture4Alt9}.
%

\begin{comment}

\section{The Design, Intent, and \\ Limitations of the Templates}
The templates are intended to {\bf{approximate the final look and page length of the articles/papers}}. {\bf{They are NOT intended to be the final produced work that is displayed in print or on IEEEXplore\textsuperscript{\textregistered}}}. They will help to give the authors an approximation of the number of pages that will be in the final version. The structure of the \LaTeX\ files, as designed, enable easy conversion to XML for the composition systems used by the IEEE. The XML files are used to produce the final print/IEEEXplore pdf and then converted to HTML for IEEEXplore.

  \section{Where to Get \LaTeX \ Help --- User Groups}
The following online groups are helpful to beginning and experienced \LaTeX\ users. A search through their archives can provide many answers to common questions.
\begin{list}{}{}
\item{\url{http://www.latex-community.org/}} 
\item{\url{https://tex.stackexchange.com/} }
\end{list}  
\end{comment}
\vspace{-2mm}
\section{Overview}
\label{sec:Overview}
\begin{figure}[t]
\centerline{\includegraphics[width=.89\linewidth]{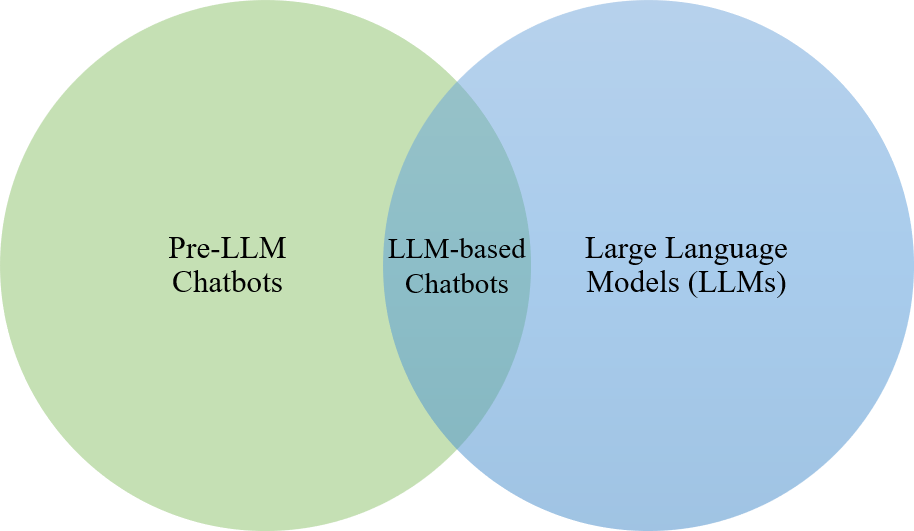}}
\caption{Pre-LLM chatbots meet LLMs.}
\label{fig:Picture72Alt}
\vspace{-6.15mm}
\end{figure}

In this section, we look into the evolution of chatbots from their origins to the present-day era. The Venn diagram in Fig. \ref{fig:Picture72Alt} illustrates the relationship between early chatbots, the development of LLMs, and the LLM-based chatbots that represent the forefront of this technology. We start by looking at the \textit{Pre-LLM Era of Chatbots} to understand the early developments in the field. Next, we introduce the \textit{Large Language Models (LLMs)}, explaining their transformative impact on chatbot technology. Finally, we provide an overview of the \textit{LLM-based Chatbots}, highlighting the frontrunners in the industry and the ones currently in development.
\vspace{-3mm}
\subsection{Pre-LLM Era of Chatbots}
\begin{figure*}[htbp]
\centerline{\includegraphics[width=.89\linewidth]{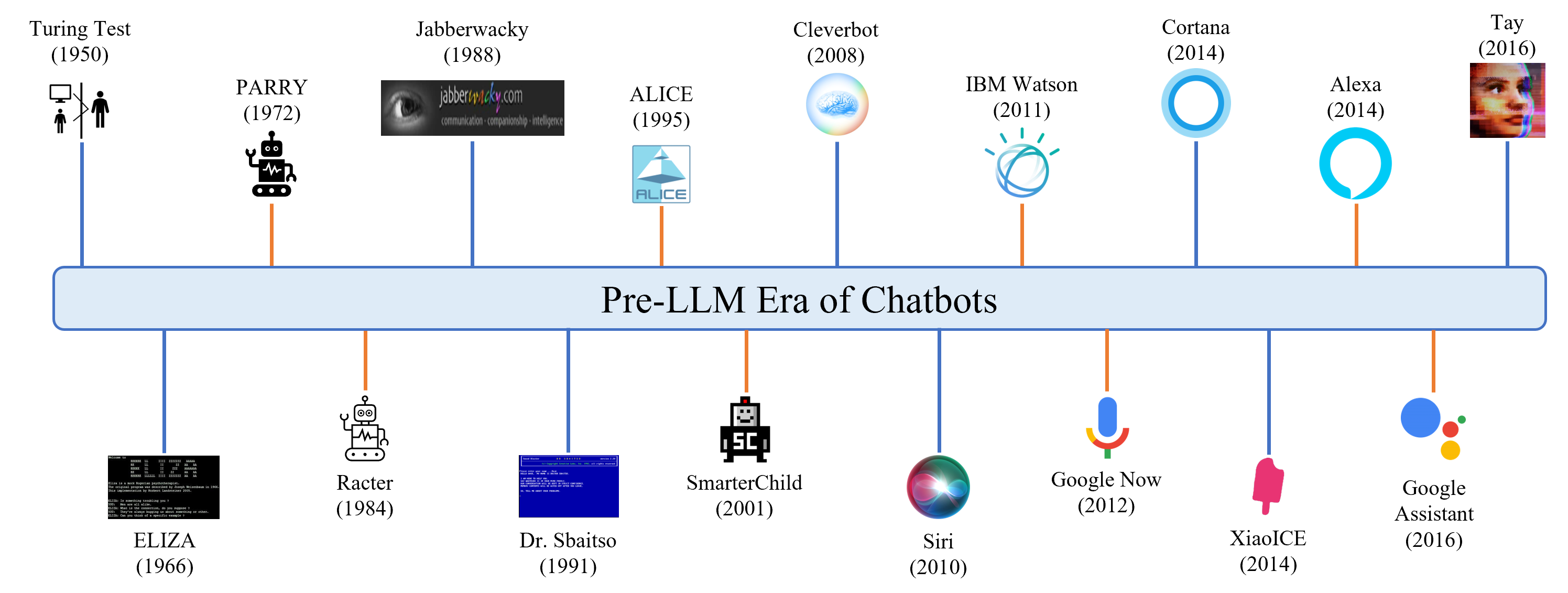}}
\caption{Pre-LLM chatbots.}
\label{fig:Picture82}
\vspace{-4.2mm}
\end{figure*}
Chatbots started with a simple idea back in 1950: “Can machines think?” This idea, called the Turing Test, was introduced by Alan Turing \cite{xue2023bias, turing2009computing}. It is where a human participant engages in textual conversation with an unseen individual and tries to predict whether that individual is a machine or another human being. This served as the foundation for chatbots. Over the years, chatbots have changed a lot, and nowadays, they use complex, advanced computer programs called large language models (LLMs) to interact with us in a more sophisticated manner. Now, in this subsection, we first discuss the era of chatbots before the arrival of LLMs.\\
\indent\textbf{1960–1980: The Early Foundations.} Several chatbots came into existence during this period. One of the first ones was ELIZA, made in 1966 at MIT. It simulated a Rogerian psychotherapist and functioned through keyword identification with pattern matching, yet it did not understand the meanings of inputs \cite{weizenbaum1966eliza}. Despite its rudimentary functionality and limited knowledge base, ELIZA gained attention for convincing users of its human-like qualities and even forming emotional bonds, which later raised some ethical considerations as well \cite{zimmerman2023human}. PARRY, another early chatbot created in 1972, was designed to simulate a person with paranoid schizophrenia \cite{colby1971artificial}. It could even chat with ELIZA and was seen as a step forward because it could exhibit a more controlled structure and emotional responses \cite{colby1981modeling, zemvcik2019brief}.\\
\indent\textbf{1981–2009: Advancements and Mainstream Integration.} During this period, chatbots became more advanced. In 1984, Racter emerged as an AI program that could produce English prose and imitate chatbot-like conversational behavior \cite{Racter}. Meanwhile, another AI project, named Jabberwacky, was initiated in 1988. It was designed to mimic casual human conversation in a friendly manner \cite{shawar2007fostering}. It evolved through interactions with humans, stored key phrases from dialogues to enhance its knowledge base, and then employed a context-aware algorithm from the dynamically expanding database to select relevant replies \cite{kerlyl2006bringing, singh2020survey}. The 1990s brought further innovations with Creative Labs' Dr. Sbaitso, a chatbot designed for MS-DOS computers. Paired with several sound cards of the era, it offered a simple interactive interface featuring a blue backdrop and white text. However, it was innovative in utilizing early text-to-speech technology through speech synthesis and sound cards \cite{deryugina2010chatterbots}. Then, in 1995, Dr. Richard S. Wallace, an American scientist, created A.L.I.C.E. (Artificial Language Internet Computer Entity), also known as Alicebot, or simply Alice. It brought new capabilities to chatbot technology by utilizing a vastly expanded knowledge base and employing Artificial Intelligence Markup Language (AIML) to establish chat guidelines \cite{wallace2009anatomy}. ELIZA was the inspiration behind the development of Alice. On its debut, Alice gained enormous appreciation for its capabilities and was awarded the Loebner Prize three times in the 2000s \cite{bradevsko2012survey}. However, it fell short of passing the Turing test due to certain constraints  \cite{shum2018eliza}. Building on this foundation, 2001 witnessed another major advancement when ActiveBuddy launched SmarterChild on the AIM platform. It was one of the first chatbots that assisted users with everyday tasks like weather updates and checking stock prices through interactive communication \cite{adamopoulou2020chatbots}. Continuing the evolution of chatbots, 2008 saw the introduction of Cleverbot by British AI scientist Rollo Carpenter. It is the successor to the 1988 chatbot, Jabberwacky. Cleverbot's unique strategy of learning from human inputs rather than relying on pre-programmed responses gives it a distinctive edge over other chatbots. Also, Cleverbot demonstrated a remarkable performance in a formal Turing test at the 2011 Techniche festival, where it earned a 59.3\% human-like rating, a noteworthy result considering that human participants scored a slightly better 63.3\% \cite{Clever}.\\
\indent\textbf{2010–2016: The Era of Intelligent Voice Assistants.} In 2011, IBM unveiled Watson, a conversational AI that won the championship title twice in the Jeopardy quiz show. Following its success, Watson found a lot of applications in the healthcare industry \cite{chen2016ibm, high2012era}. Then, in 2014, Microsoft launched XiaoICE \cite{zhou2020design}. This chatbot, built on an emotional computing framework, could handle queries with both intellectual and emotional quotients. The same team from Microsoft created another chatbot named Tay. Tay made its first appearance on Twitter in 2016. However, soon after its launch, Tay began to post offensive tweets that forced Microsoft to shut it down within just sixteen hours after its release. During this period, the integration of chatbots into everyday tasks became more prominent through voice and search agents on instant messaging apps and various platforms \cite{hoy2018alexa, kepuska2018next}. Apple pioneered this integration in 2010 with Siri, an iOS app that became part of the iOS system by 2011. As a personal assistant, Siri can execute a range of tasks, such as making calls, setting reminders, and gathering information, all through voice commands \cite{aron2011innovative}. Subsequently, in 2012, Google launched Google Now, which transformed voice inputs into search results. Then Microsoft followed in 2014 with Cortana for Windows, which utilized Bing for user queries. That same year, Amazon released Alexa alongside the Echo speaker. Soon after, in 2016, Google further advanced the field with Google Assistant, which later got integrated into Google Home speakers and Pixel smartphones. Although these voice assistants provide swift internet-connected responses, they have several issues related to multilingual support, privacy, and security \cite{bolton2021security}.\\
% Table Start
\def\arraystretch{1.21}
\begin{table}
\centering
\captionsetup{labelsep=newline,justification=centering}
\caption{Overview of the LLMs}
\label{tab:Table1}
\resizebox{\linewidth}{!}{%
\begin{tabular}{|c|c|c|c|} 
\hline
& Dataset & Parameters & Context Window \\
\hline
GPT-1 & \begin{tabular}[c]{@{}c@{}}BooksCorpus\\(4.5GB)\end{tabular} & 117 million & 512 tokens \\ 
\hline
BERT & \begin{tabular}[c]{@{}c@{}}BooksCorpus,\\ English Wikipedia\\ (Size: N/A)\end{tabular} & \begin{tabular}[c]{@{}c@{}}BERT-Base: 110 million\\ BERT-Large: 340 million\end{tabular} & 512 tokens \\ 
\hline
GPT-2 & \begin{tabular}[c]{@{}c@{}}Webtext\\(40GB)\end{tabular} & 1.5 billion & 1024 tokens \\ 
\hline
GPT-3 & \begin{tabular}[c]{@{}c@{}}Common Crawl\\(45TB)\end{tabular} & 175 billion & 2048 tokens \\ 
\hline
GPT-3.5 & N/A & N/A & 2048 tokens \\ 
\hline
PaLM & \begin{tabular}[c]{@{}c@{}}Webpages, Books, News, \\ Social Media Conversations, \\ Wikipedia, Github \\ (Size: N/A)\end{tabular} & \begin{tabular}[c]{@{}c@{}}540 billion \\ Smaller Versions: \\ 8 million and 62 billion\end{tabular} & N/A \\ 
\hline
LLaMA & \begin{tabular}[c]{@{}c@{}}Common Crawl, C4, \\ Books, Github, Wikipedia \\ ArXiv, Stack Exchange \\ (Size: N/A)\end{tabular} & \begin{tabular}[c]{@{}c@{}}6.7, 13, 32.5, \\ and 65.2 billion\end{tabular} & 2048 tokens \\
\hline
GPT-4 & N/A & N/A & 8195 tokens \\ 
\hline
PaLM 2 & \begin{tabular}[c]{@{}c@{}}20 Programming, \\ Languages, Over 100 \\ spoken Languages, Math \\ and Science texts \\ (Size: N/A)\end{tabular} & 340 billion & 8000 tokens \\
\hline
LLaMA 2 & \begin{tabular}[c]{@{}c@{}}Mix of Publicly \\ Available Online Sources \\(Size: N/A)\end{tabular} & \begin{tabular}[c]{@{}c@{}}7, 13, and \\ 70 billion\end{tabular} & 4096 tokens \\
\hline
\end{tabular}
}
\vspace{-4mm}
\end{table}
\indent Fig. \ref{fig:Picture82} provides a complete timeline of all the chatbots mentioned so far. Now, regardless of all these advancements throughout the decades, the new era of chatbots began in 2020 with the rise of LLMs \cite{wei2023overview, zhou2023comprehensive}. Through their extensive training on vast pre-trained transformers, LLMs have empowered chatbots to deliver more detailed and nuanced responses. The following sections will discuss these LLMs and popular AI-driven chatbots built on this technology.
% Large Language Models (LLMs)
\subsection{Large Language Models (LLMs)}
\begin{figure}[t]
\centerline{\includegraphics[width=\linewidth]{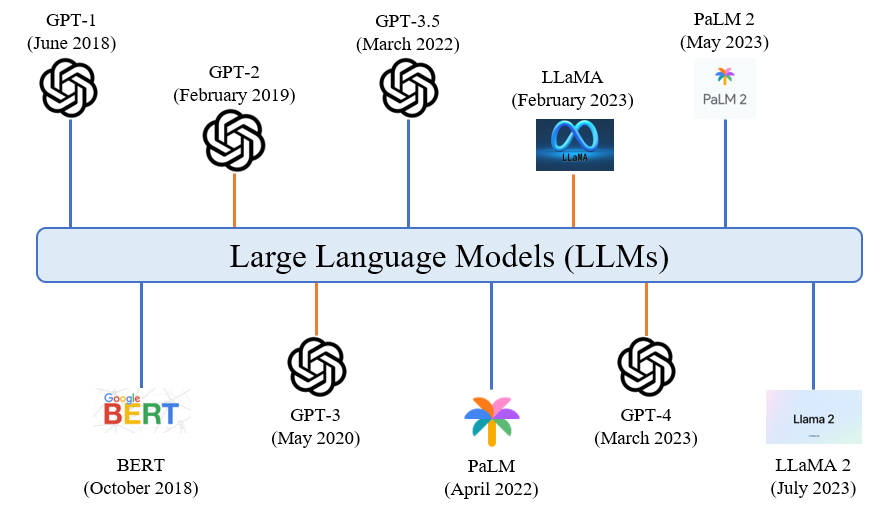}}
\caption{Timeline of LLMs.}
\label{fig:Picture95}
\vspace{-5mm}
\end{figure}

\begin{figure}[htbp]
\centerline{\includegraphics[width=0.95\linewidth]{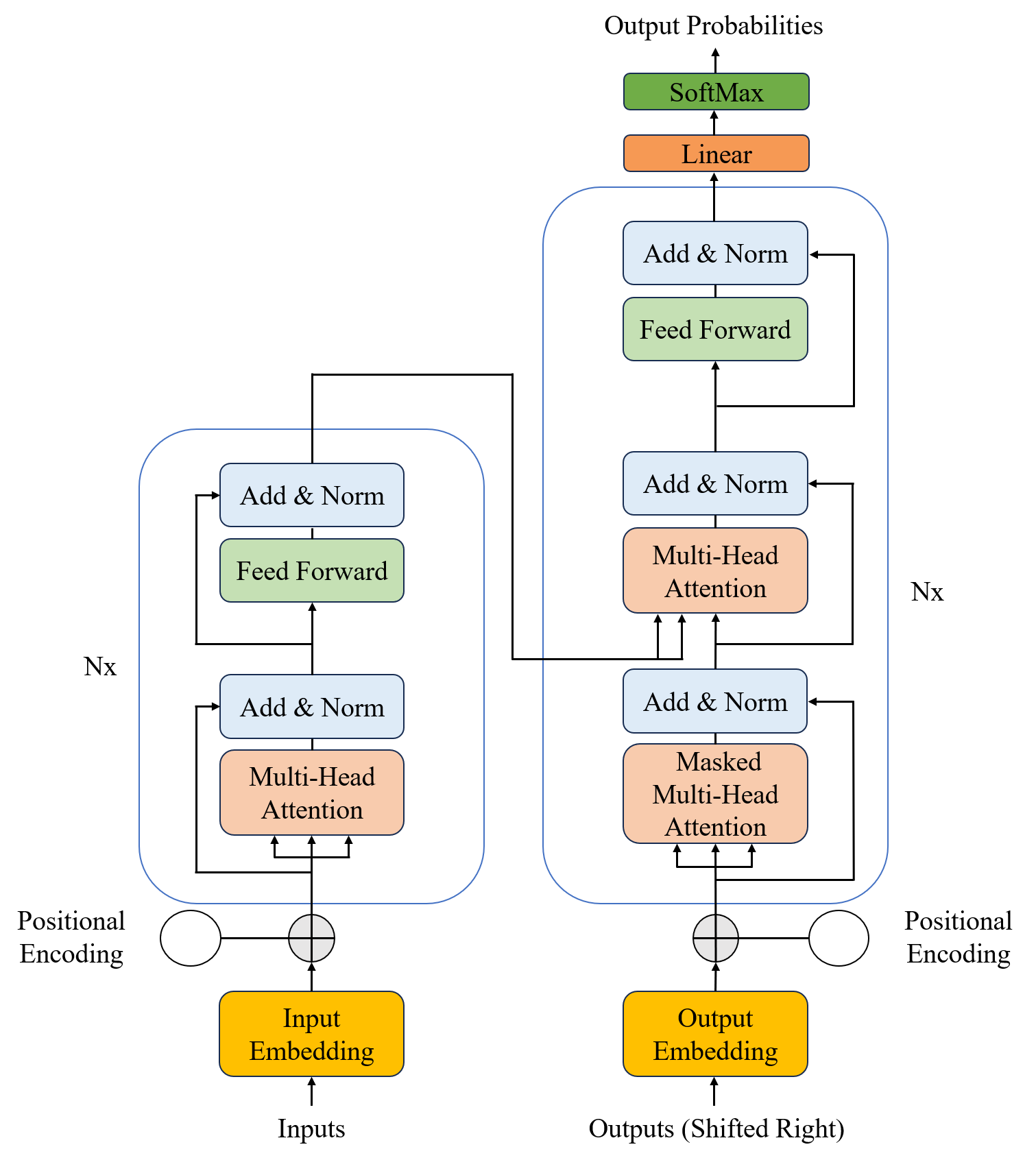}}
\caption{Transformer model architecture \cite{vaswani2017attention}.}
\label{fig:Picture1Alt}
\vspace{-5mm}
\end{figure}

\begin{figure*}[t]
\centerline{\includegraphics[width=.9\linewidth]{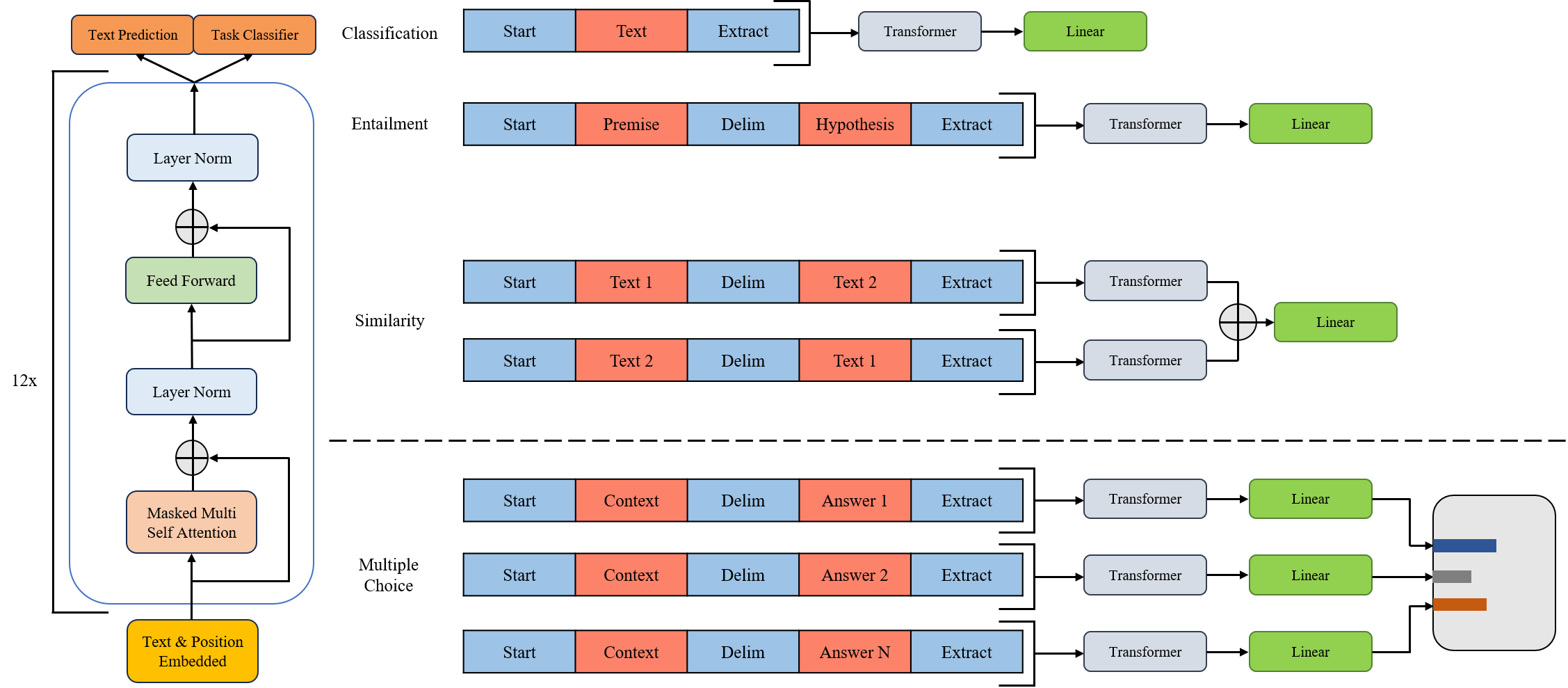}}
\caption{Transformer model architecture in the GPT series \cite{radford2018improving}.}
\label{fig:Picture3}
\vspace{-5mm}
\end{figure*}

The advent of LLMs has revolutionized the domain of natural language processing, particularly the development and functionality of chatbots. Here, we discuss the world of LLMs, providing an overview of their architecture, working principles, applications in chatbots, advantages, and limitations.\\
% Then, we talk about the LLM-based chatbots already in use and those in development.\\
\indent LLM-based chatbots like ChatGPT and BARD have recently gained widespread acclaim from the media \cite{JournalistGPT, JournalistBARD, JournalistGPTandBARD}, policymakers \cite{Popularity, Risk, AIFreaks}, and academics across many sectors \cite{ray2023chatgpt, sallam2023chatgpt, choi2023chatgpt, kitamura2023chatgpt}. LLMs, often referred to as transformer language models, are trained on enormous amounts of text data and include billions of parameters. The first LLM to draw people's attention was OpenAI's GPT \cite{radford2018improving}, the Generative Pre-trained Transformer, released in 2018. Since then, we've witnessed the development of larger, more sophisticated language models, including GPT-2 \cite{radford2019language}, GPT-3 \cite{brown2020language}, GPT-3.5 \cite{ye2023comprehensive}, and the most recent GPT-4 \cite{openai2023gpt4}, along with other models like BERT \cite{devlin2018bert}, PaLM \cite{chowdhery2022palm}, and LLaMA \cite{touvron2023llama}. Table \ref{tab:Table1} provides an overview of these LLMs, while Fig. \ref{fig:Picture95} shows their development timeline. Currently, encoder-decoder, causal decoder, and prefix decoder are the three basic types of transformer architecture upon which the LLMs are constructed.\\
%\vspace{-3.2mm}
\indent The vanilla transformer model, proposed by Vaswani et al. \cite{vaswani2017attention} in their paper titled “Attention is All You Need,” is built on the encoder-decoder architecture. The encoder produces abstract representations of the input data and a series of layers with multi-head self-attention to focus on different parts of the input sequence.
The decoder then autoregressively generates the output sequence, using cross-attention on these representations (see Fig. \ref{fig:Picture1Alt}). The GPT-series models \cite{brown2020language, wolf2020transformers} use an autoregressive or causal decoder architecture with a one-way attention mask that lets each input token think about only the elements that came before it and itself while it is being processed (see Fig. \ref{fig:Picture3}). This makes the processing more like how a conversation would naturally flow. Both input and output tokens undergo similar processing within this framework.
The non-causal or prefix decoder architecture \cite{zhang2022examining} performs bidirectional attention on prefix tokens, i.e., it considers both preceding and subsequent tokens. While autoregressively predicting the output tokens utilizing the same set of parameters used in encoding, it performs unidirectional attention \cite{liu2023pre, dong2019unified}.\\
\begin{comment}
\begin{figure}[htbp]
\centerline{\includegraphics[width=0.95\linewidth]{Picture2Alt.png}}
\caption{Self-attention module in transformer architecture.}
\label{fig:Picture2}
\vspace{-3mm}
\end{figure}    
\end{comment}
\indent The working principle of LLMs includes a series of steps. The process begins with word embedding, which involves representing words as vectors in a high-dimensional space. Here, similar words are clustered in specific groups or categories. This clustering of words enables the model to understand their meanings, which helps the LLM make accurate predictions. The model is trained on large text corpora like news articles or books, and during training, it learns to predict the likelihood of a word appearing in a specific context. Positional encoding further adds an understanding of word order within sequences, essential for tasks like translation, summarization, and question answering. Then comes the core of these models, the transformer architecture. It consists of the self-attention mechanism%
that helps understand textual dependencies by allocating distinct weights to individual words. This is achieved by computing:
\begin{equation}
Attention(Q, K, V) = softmax(\frac{QK^T}{\sqrt{d_k}})V = AV,
\end{equation}
where matrices \textbf{\textit{Q}} (\textit{query}), \textbf{\textit{K}} (\textit{key}), and \textbf{\textit{V}} (\textit{value}) represent the current element, other elements, and information to be aggregated, respectively. The similarity between the query and key matrices is calculated through a dot product operation. It is then scaled by $\frac{1}{\sqrt{d_k}}$ to prevent gradient vanishing problems and subsequently normalized using the \textbf{\textit{SoftMax}} activation function to produce the attention matrix \textbf{\textit{A}}. The updated representations are obtained by performing matrix multiplication on \textbf{\textit{A}} and \textbf{\textit{V}}. This aggregation of weighted values forms a new representation that captures the inherent associations within the text. Algorithm \ref{alg:your_alg} outlines the step-by-step procedure for this self-attention mechanism. Lastly, the model generates contextually relevant text in response to a given prompt using an autoregressive approach, where the model builds the output sequence one word at a time. The integration of reinforcement learning from human feedback (RLHF) further empowers LLMs to learn from human interactions, continually refining their performance.\\
\begin{algorithm}[b]
\caption{Self-Attention Mechanism}
\label{alg:your_alg}
\begin{algorithmic}[1] % The [1] enables line numbers
\REQUIRE Matrices $Q$ (Query), $K$ (Key), $V$ (Value)
\ENSURE Matrix $Z$ (Updated Representations)
\STATE Compute dot products of queries with keys: $D \gets QK^T$
\STATE Scale dot products: $D \gets D / \sqrt{d_k}$
\STATE Apply SoftMax normalization: $A \gets \text{SoftMax}(D)$
\STATE Compute the weighted sum of values: $Z \gets AV$
\RETURN $Z$
\end{algorithmic}
\end{algorithm}
% Table End
\begin{comment}
\begin{table*}
\centering
\caption{Capabilities of AI in Educational Support}
\label{table:ai_education_support}
\begin{tblr}{
  width = \linewidth,
  colspec = {Q[129]Q[167]Q[644]},
  %cells = {black,c},
  cell{2}{1} = {r=3}{},
  cell{5}{1} = {r=2}{},
  hline{1-2,5,7} = {-}{},
  hline{3-4,6} = {2-3}{},
}
Aspect                & Function                       & Representative Quotes                                                                                                                       \\
Teaching Preparation  & Creating course materials      & “We requested ChatGPT to format dialogues for DialogFlow integration, and it successfully generated the required format.” [42, p. 37].      \\
                      & Giving suggestions             & “When informed about a learner with dyslexia, ChatGPT was able to propose suitable learning resources.” [41, p. 1].                         \\
                      & Providing language translation & “ChatGPT has the capacity to adapt educational content for various linguistic audiences.” [46, p. 8].                                       \\
Helping with Assessment & Crafting assessment tasks      & “A notable application of ChatGPT lies in its creation of practical exercises and tests for classroom use.” [29, p. 5].                     \\
                      & Grading academic performance   & “ChatGPT is adept at evaluating student essays, which allows educators to devote more attention to other instructional duties.” [29, p. 8]. 
\end{tblr}
\end{table*}
\end{comment}
%\vspace{-1mm}
\indent Introduced in GPT-3 \cite{brown2020language}, the in-context learning (ICL) feature of the LLMs allows them to understand and respond to new information in a conversational context without any additional training. LLMs can adhere to the instructions of the input text and produce output that complies with those instructions as much as possible. By instruction tuning (a fine-tuning process), LLMs are further trained on a blend of multi-task datasets, each shaped with natural language instructions, which enhances the model's generalization ability to perform well on unfamiliar tasks described using similar instructions \cite{wei2021finetuned, sanh2021multitask, ouyang2022training}. Unlike small language models, LLMs can handle complex tasks involving multiple reasoning steps by employing a strategy known as chain-of-thought (CoT) prompting \cite{wei2022chain}. This strategy helps the LLMs outline the intermediate steps necessary to reach the final solution. In other words, instead of jumping from the problem to the solution in one step, CoT prompting breaks down the task into several parts that the LLMs can solve sequentially, leading to the solution.\\
\indent Although LLMs effectively generate coherent text, they lack semantic understanding. It is because they do not genuinely understand the content. They only predict the subsequent text based on what they have learned from the training data. LLMs can also process and generate multilingual text, given sufficient multilingual training data. However, proficiency varies with data quality and quantity for different languages. There are some other limitations as well. For example, they occasionally create hallucinations \cite{openai2023gpt4, bang2023multitask, wu2023brief}, where the responses contain factual errors or can be considered risky in some situations. While producing text with complex structural constraints, LLMs demonstrate excellent proficiency in maintaining local planning. That is, they can efficiently handle interactions between closely adjacent sentences. They can, however, struggle with global planning or maintaining coherence and relevance over longer stretches of text \cite{bubeck2023sparks}.\\
\indent To conclude, LLMs have transformed the landscape of natural language processing by providing strong capabilities for comprehending and generating human-like text. Despite their remarkable advancements, there are still some limitations. To ensure their ethical and appropriate applications across diverse sectors, we must improve them as we move forward.
\begin{comment}
\section{Other Resources}
See \cite{ref1,ref2,ref3,ref4,ref5} for resources on formatting math into text and additional help in working with \LaTeX .    
\end{comment}
\vspace{-2mm}
\subsection{LLM-based Chatbots}
The development of advanced chatbots has been made possible by LLMs. Today's market offers a wide range of chatbots, with ChatGPT leading from the front. Here, we discuss the chatbots that are now in use and those currently in the development phase.\\
\indent\textbf{ChatGPT.} In November 2022, we saw the emergence of ChatGPT, an AI chatbot developed by OpenAI \cite{ChatGPTFirst}. It belongs to the larger family of Generative Pre-trained Transformers (GPT) and is specifically a fine-tuned version of GPT-3.5 \cite{zhang2023one}. Leveraging a large corpus of internet text data for its training, ChatGPT can generate human-like replies to numerous prompts and inquiries. Within a relatively short time, ChatGPT gained widespread acclaim for its ability to deliver coherent and convincingly realistic responses on several topics. Following the unprecedented success of ChatGPT, OpenAI released GPT-4 on March 14 last year \cite{ali2022performance}. GPT-4 is the fourth and latest installment in the GPT series, as well as the underlying architecture of ChatGPT Plus.\\
\indent\textbf{BARD.} Since its release, ChatGPT has made a formidable impact on search engines, to the extent that Google declared a ‘code red’ in response to its emergence \cite{zhang2023one}. Acknowledging ChatGPT's potential, Google unveiled BARD, a public-access user interface to an LLM for collaborative generative AI. BARD was launched on February 6 and made publicly accessible on March 21 of last year \cite{GoogleBARDWiki}. It uses an optimized variant of LaMDA (Language Models for Dialogue Applications), pre-trained on a broad spectrum of publicly available resources. At the time of its release, one of the main differences between ChatGPT and BARD was that ChatGPT's responses could not use up-to-date information since its knowledge was confined to data available up to 2021, whereas BARD leverages more recent information \cite{BARDvsChatGPT}. However, since September of last year, ChatGPT has also gained the ability to search the web for the most recent content \cite{ChatGPTUpdate}, enhanced by an updated knowledge base that includes information up to April 2023 \cite{ChatGPTKnowledge}. Nevertheless, Google has further addressed areas such as accuracy, bias, and vulnerability for continuous research and improvement over time. It is important to note that, as of February 8, 2024, Google has rebranded BARD to ‘Gemini’ \cite{GoogleBardBecomeGemini}. This survey primarily references publications from 2023, when the platform was universally recognized as BARD. Hence, all mentions of BARD within this document refer to the chatbot now known as Gemini.\\
\indent\textbf{Bing Chat.} On February 7 last year, shortly after Google announced BARD, Microsoft introduced Bing Chat \cite{BingChatAll}. It is a search engine feature powered by GPT-4 that lets users engage with an AI chatbot rather than manually typing search queries. Upon its release, Bing Chat held a significant edge over competitors like ChatGPT, as it provided live internet access and citation-supported responses, enabling users to validate the authenticity of the information \cite{BingChatCitation, BingChatCitation2}. Notably, Bing Chat differentiates itself further by offering user-selectable response styles like ‘More Creative,’ ‘More Balanced,’ or ‘More Precise,’ providing a customized interaction based on the user's query \cite{BingChatCustomization}.\\
\indent\textbf{Claude.} The first iteration of Anthropic's Claude, version 1.0, was launched on March 14 last year, alongside a streamlined version, Claude Instant 1.1 \cite{ClaudeOrigin}. It was later followed by the more advanced Claude 2 on July 11 and the second iteration of the Claude Instant, version 1.2, on August 9 \cite{ClaudeDates}. Then Claude 2.1 came out on November 21 \cite{ClaudeDates}. Although Claude Instant is a bit faster and lighter than the other two models, Claude 2 stands out for its comprehensive reasoning abilities and safer responses. It is achieved through constitutional AI, a fine-tuning process developed by Anthropic researchers \cite{lozic2023chatgpt}. However, the knowledge base of Claude 2 contains data up to 2022, and it cannot connect to the internet. Therefore, it lacks real-time data after that period [86]. Another key feature of Claude 2 is its extensive context window of 100,000 tokens, or about 75,000 words, that allows users to conduct an in-depth analysis of large documents \cite{lozic2023chatgpt, bai2022constitutional, bai2022training}.\\
\indent\textbf{Ernie Bot.} Baidu's Ernie (Enhanced representation through knowledge integration) was first unveiled on March 16 last year \cite{motlagh2023impact}. It is also known as Wenxin Yiyan (language and mind as one). Ernie is trained on a vast array of data, including trillions of web pages, billions of voice data, search data, image data, and a knowledge graph of 550 billion facts, which is an impressive feat \cite{yang2023chinese}. Despite some early testing issues with hallucinations and elementary math errors, Ernie has shown the capability to read texts in various Chinese dialects \cite{rudolph2023war}. Baidu has plans to integrate Ernie into its numerous products, including autonomous vehicles and its primary search engine \cite{yang2023chinese}. Baidu's strategy might not aim to compete directly with ChatGPT but rather to establish domestic market dominance where ChatGPT is currently unavailable \cite{ErnieinsteadChatGPT}.\\
\begin{figure}[t]
\centerline{\includegraphics[width=0.85\linewidth]{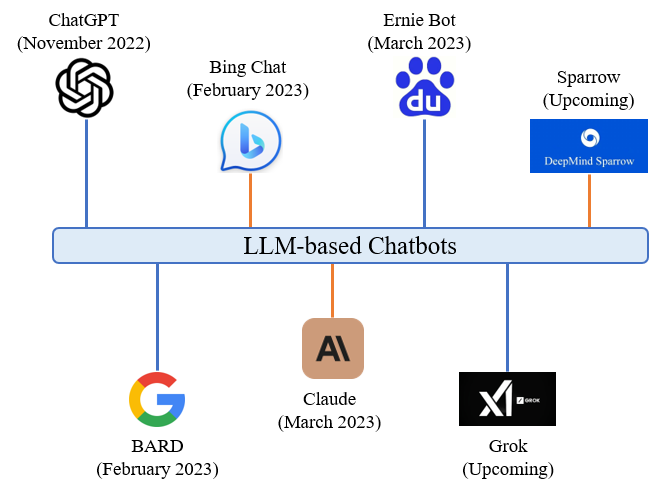}}
\caption{Timeline of LLM-based chatbots.}
\label{fig:Picture105}
\vspace{-5mm}
\end{figure}
\indent In addition to these five, several other chatbots are also under development, with promising candidates like DeepMind's Sparrow \cite{Sparrow} and xAI's Grok \cite{Gork, GorkTechcrunch}. Fig. \ref{fig:Picture105} provides a visual timeline of all these chatbots. There is also one chatbot from Meta known as BlenderBot \cite{zhang2023one}. However, the users have found the interactions with the BlenderBot somewhat lackluster, possibly because of the tight constraints on its output set by developers, limiting the chatbot's creativity and diminishing user satisfaction \cite{BlenderBotIssues}. Nevertheless, as we witness this rapid progress, the significance of continued research, cross-disciplinary collaboration, and ethically sound development practices becomes paramount. Balancing innovation with ethical considerations will be crucial in ensuring the responsible utilization of these tools.
\vspace{-2mm}
\section{Applications}
\label{sec:Applications}
Unlike traditional chatbots, which are limited to basic conversational frameworks, LLM-based chatbots have become a new way to generate knowledge. Due to this new role, they have emerged as integral components across various domains, reshaping how industries operate and interact with their customers. In sectors like education, research, and healthcare, these chatbots are providing great efficiency, accuracy, and personalized engagement. Now, in this section, we explore the diverse applications of LLM-based chatbots, highlighting their profound impact across different sectors.
\vspace{-2mm}
\subsection{Education}
LLM-based chatbots bring considerable enhancements to the realm of education. Here, we summarize how these intelligent systems offer unique opportunities for educational excellence and improved learning outcomes.\\
\indent\textbf{Learning.} LLM-based chatbots hold great potential across various levels of education, where they could play a range of supportive roles at K–12, undergraduate, and graduate levels, enhancing the learning experience across these educational tiers. For example, a study in \cite{vasconcelos2023enhancing} explores the promising avenues of ChatGPT and Bing Chat in STEM education. The study shows that these chatbots, as ‘objects-to-think-with,’ can transform STEM education, encourage active participation, and foster user-friendly environments. Following this, another study \cite{dao2023large} investigates the performance of the three SOTA LLMs, ChatGPT, Bing Chat, and BARD, in meeting the educational needs of Vietnamese students. The study provides a comparative analysis of students’ academic results in nine subjects using the Vietnamese National High School Graduation Examination (VNHSGE) dataset \cite{xuan2023vnhsge}. Though all the chatbots exhibit comparable performances, Bing Chat emerges as the more advantageous option as it surpasses the other two in the majority of subjects. It achieves notable performances in Mathematics, English, Physics, History, and Civic Education, with accuracies of 60\%, 92.4\%, 66\%, 88.5\%, and 85.5\%, respectively. BARD also does reasonably well in subjects like Chemistry, Biology, and Geography, with 73\%, 69.5\%, and 82\% accuracy, respectively. In contrast, ChatGPT excels only in Literature with 68\% accuracy but has lower scores in other subjects. Apart from these, GPT-4’s meta-prompt feature allows for role customization during conversations, such as selecting a ‘Socratic tutor mode’ to enhance the critical thinking and problem-solving skills of the students \cite{thirunavukarasu2023large}. A similar approach can also be seen in ChatGPT Turbo, with features like ‘Math Mentor,’ ‘Creative Writing Coach,’ ‘Data Analysis,’ and a few more that facilitate customized learning experiences for the users \cite{ToolifyAI}.\\
% These interactive sessions help teachers to monitor their students’ progress and adjust their teaching methods accordingly \cite{thirunavukarasu2023large}
% In addition, GPT-4’s meta-prompt feature allows for role customization during conversations, such as selecting a ‘Socratic tutor mode’ to enhance the critical thinking and problem-solving skills of the students. These interactive sessions help teachers to monitor their students’ progress and adjust their teaching methods accordingly \cite{thirunavukarasu2023large}. A similar approach can also be seen in ChatGPT Turbo, with features like ‘Math Mentor,’ ‘Creative Writing Coach,’ ‘Data Analysis,’ and a few more that foster customized learning experiences for the users.
%Moving forward, these chatbots can also help students improve their reading skills. For instance, \cite{kasneci2023chatgpt} demonstrates how LLMs such as ChatGPT provide summaries for complex texts and prompt explanations for words and phrases so that students understand even the subtle meanings.\\
\indent\textbf{Academic Writing.} The 2020 NSF Science and Engineering Indicator highlights the significant contribution of international scholars in U.S. postdoctoral programs, with nearly 49\% of the fellows being from overseas. In Mathematics and Engineering, international students claim 60\% of PhD degrees \cite{ForeignStudents}. Therefore, effective academic writing is crucial to the quality and success of research publications. However, mastering academic writing remains a formidable challenge for many international postdocs and students. Addressing this, ChatGPT significantly enhances the user experience by fixing punctuation, spelling, and grammatical mistakes, which consequently improves both content quality and individual writing skills \cite{zhang2023one}. Besides, ChatGPT helps users develop a unique writing style. For example, it gives style suggestions, elevates the content depth, and crafts an engaging final product for the readers \cite{ChatGPTvsGrammarly}.\\
\indent\textbf{Teaching.} LLM-based chatbots can also serve as an assistant for instructors. It is noted in \cite{tlili2023if} that ChatGPT can be a valuable tool for instructors, aiding them in crafting their curriculum by providing structured outlines. Research in \cite{lo2023impact} identifies ChatGPT’s five key roles in two primary categories: aiding in teaching preparation (like creating course materials, giving suggestions, and providing language translation) and helping with assessment (such as crafting assessment tasks and grading academic performance). Table \ref{tab:Table2} illustrates ChatGPT’s utility in supporting instructors with their teaching tasks. Another study in \cite{megahed2023generative} shows that ChatGPT effectively creates an undergraduate syllabus for a Statistics course that requires only slight adjustments, which further proves its usefulness and precision. Following this trend, Khan Academy, a well-known online education platform, is currently exploring ways to integrate AI tools like GPT-4 in their ‘Khanmingo’ project to improve e-learning \cite{khan2023harnessing}, while Duolingo is already using GPT-4 to enhance the interactivity of role-play in language learning \cite{team2023introducing}.\\
\begin{table*}[!ht]
\centering
\captionsetup{labelsep=newline,justification=centering}
\caption{Capabilities of AI in Educational Support}
\label{tab:Table2}
\begin{tblr}{
  width = \textwidth,
  colspec = {|Q[c,m,0.14\linewidth]|Q[c,m,0.19\linewidth]|Q[c,m,0.42\linewidth]|},
  row{1} = {font=\bfseries, c},
  row{1-6} = {font=\linespread{1.1}\selectfont}, % Adjust line spread for all cells, if needed
  hlines,
  cell{2}{1} = {r=3}{halign=c,valign=m},
  cell{5}{1} = {r=2}{halign=c,valign=m},
}
Aspect                & Function                       & Representative Quote\\
Aiding in Teaching Preparation  & Creating Course Materials      & ‘We requested ChatGPT to format dialogues for DialogFlow integration, and it successfully gave the required format’ \cite{topsakal2022framework}.\\
                      & Giving Suggestions             & ‘When informed about a learner with dyslexia, ChatGPT recommended suitable learning resources’ \cite{zhai2023chatgpt}.\\
                      & Providing Language Translation & ‘ChatGPT is capable of translating educational content into different languages’ \cite{baidoo2023education}.\\
Helping with Assessment & Crafting Assessment Tasks      & ‘A notable application of ChatGPT lies in its ability to generate practical exercises and tests for classroom use’ \cite{wang2023chatgptMDPI}.\\
                      & Grading Academic Performance   & ‘ChatGPT can evaluate student essays, which allows teachers to give more attention to other responsibilities’ \cite{wang2023chatgptMDPI}.
\end{tblr}
\vspace{-0.20mm}
\end{table*}
\indent In addition to what we have discussed so far, LLM-based chatbots are transforming education through numerous other innovative approaches. For example, they offer grammar exercises, interactive discussions, instant feedback, and aid in translation, thereby enhancing language fluency and understanding. They can also help students improve their reading skills. \cite{kasneci2023chatgpt} demonstrates how LLMs can assist with summaries for complex texts and prompt explanations for words and phrases so that students understand even the subtle meanings. Besides, the integration of these chatbots with speech-to-text and text-to-speech tools has the potential to benefit visually impaired learners \cite{kasneci2023chatgpt}. To conclude, LLM-based chatbots represent a significant milestone in educational development, offering diverse and innovative solutions that enhance learning experiences, cater to individual needs, and pave the way for a more dynamic learning environment.
\vspace{-2mm}
\subsection{Research}
The following discussion explores how LLM-based chatbots are opening new avenues for academic research, ranging from literature reviews and paraphrasing to advanced data analysis techniques and enhanced idea-generation processes.\\
\indent\textbf{Literature Review and Paraphrasing.} A comprehensive literature review can be a time-consuming and energy-draining task for researchers. For instance, the AI-driven Semantic Scholar search engine has indexed a staggering 200 million scholarly publications. In such a vast ocean of information, finding relevant research papers and extracting key insights may seem like searching for a needle in a haystack. ChatGPT streamlines the exploration of numerous papers by finding related literature on a given topic \cite{chandha2023setting}. Besides, similar to SciSpace Copilot, ChatGPT can also explain scientific literature and mathematics in multiple languages, which facilitates a better understanding of the research article \cite{chandha2023setting}. Beyond this, ChatGPT, as a versatile language model, extends its utility across various other natural language processing tasks. For example, in a recent study \cite{aydin2022openai}, ChatGPT has shown promising results in paraphrasing abstracts related to ‘Digital Twin in Healthcare.’  However, its application in the literature review is in its initial stages. While ChatGPT’s literature review capabilities are evolving, its potential to enhance researchers’ efficiency and focus on core research is anticipated. Another recent paper \cite{aydin2023google} explores the use of Google BARD in generating literature reviews.  The author collects ten Metaverse articles published in recent years (2021–2023) from Google Scholar and uses BARD to paraphrase their abstracts. Subsequently, the author asks Google BARD, “What is the Metaverse?” All the texts are then scrutinized using the iThenticate plagiarism checker. While the results are promising, the author observes that the paraphrased texts exhibit a 12\% plagiarism-matching rate, which is considerably higher than the 1\% rate observed in BARD's response to the Metaverse query. Nevertheless, this experiment underscores the potential of LLM-based chatbots, indicating their growing significance in academic research.\\
\indent\textbf{Data Analysis.} The process of preparing and organizing scientific data for analysis can be a time-consuming task, often stretching over months. In addition, researchers also need to acquire coding skills such as Python or R. The integration of ChatGPT into data processing offers a transformative shift, enhancing research efficiency and methodology. For instance, a study in \cite{macdonald2023can} shows that ChatGPT efficiently handles a simulated dataset featuring 100,000 healthcare workers with varied ages and risk profiles. Another recent article \cite{ChatGPTData} demonstrates that LLM-based chatbots, such as ChatGPT 3.5 and 4, can effectively handle fundamental data summarization tasks using the Pandas DataFrame Agent. These models can answer essential exploratory data analysis (EDA) questions, such as identifying the date with the highest average price or determining the correlation between two variables. However, they sometimes struggle to generate meaningful context, like producing overlapping periods that do not make sense. Despite these limitations, ChatGPT 4 shows significant promise in generating valuable insights on investment opportunities, risks, and services through prompt engineering and value-chain analysis workflows. The challenge arises when users upload extensive data, such as earnings calls and annual reports, leading to data flow and latency issues that often cause runtime errors. A practical solution involves dividing documents into segments and using a memory buffer to compile responses. Besides, integrating textual and numerical data from multiple sources through custom agents can enhance the chatbot's contextual understanding, offering more thoughtful insights.\\
\indent\textbf{Idea Generation.} One of the most fundamental components of research is the ability to think critically and generate innovative ideas. LLM-based chatbots can assist students and teachers in this aspect of their research by acting as advanced tools for idea generation. \cite{girotra2023ideas} demonstrates how ChatGPT can play a vital role in research, from stimulating creativity for new idea generation to providing suggestions for expanding existing ideas. \cite{temsah2023reflection} further states how ChatGPT offers insights from multiple viewpoints as it explores the consequences of the COVID-19 pandemic by analyzing its effects from various dimensions, including the healthcare system, socio-economic implications, and individual health practices. This ability to analyze issues from different perspectives helps in generating comprehensive and well-rounded ideas.\\
\indent Before 2022, it was thought AI would be best for simple tasks, and the domain of creative works would remain in the hands of humans. While these tools are not always accurate, their unbiased approach leads to substantial improvements in various aspects of research, sometimes outperforming the average human in creativity.
\vspace{-3.05mm}
\subsection{Healthcare}
Here, we summarize how LLM-based chatbots are reshaping the healthcare domain, offering advanced support in addressing complex medical queries, patient care, and treatment suggestions.\\
\indent\textbf{Question Answering.} One of the key highlights of LLM-based chatbots is their vast knowledge base, particularly evident in automatic question-answering systems in the medical sector. For example, \cite{gilson2023does} shows ChatGPT's ability to handle questions from the United States Medical Licensing Examination (USMLE) steps 1 and 2 exams. The study then compares those responses from ChatGPT with those of InstructGPT and GPT-3, where ChatGPT outperforms InstructGPT by 8.15\% on average, while GPT-3's responses are less consistent. ChatGPT also shows competency similar to a third-year medical student's passing grade. Another study \cite{hamidi2023evaluation} evaluates Claude and ChatGPT 3.5 in answering clinical questions using MIMIC-III clinical notes (from TREC CDS 2016 topics \cite{kirk2016overview}). The study then compares those answers in terms of accuracy, coherence, relevance, and coverage. Kruskal-Wallis analysis of variance \cite{kruskal1952use} further validates the findings. Results show that both Claude and ChatGPT 3.5 effectively answer clinical questions based on admission notes, providing precise, relevant, and clear responses in various scenarios. Another recent article \cite{azizi2023evaluating} investigates how ChatGPT and Bing Chat (GPT-4) respond to patient and clinician questions on Atrial Fibrillation (AF). The authors prepare eighteen patient-centric prompts in consultation with experts in AF management and another eighteen clinician-focused prompts. Results show that ChatGPT accurately answers 83.3\% of patient queries. For the clinician-based prompts, ChatGPT and Bing Chat show text accuracies of 33.3\% and 66.6\%, with references being 55.5\% and 50\% accurate, respectively.\\
\indent\textbf{Patient Education.} In recent developments, GPT-4 and Med-PaLM 2 have demonstrated considerable effectiveness in health assessment, marking significant progress in the field of patient care technology \cite{thirunavukarasu2023large, nori2023capabilities}. A study in \cite{yang2023large} shows how LLMs provide personalized patient education with improved understanding and engagement. One recent example is Macy, an AI pharmacist. It uses ChatGPT as its underlying architecture and features a photorealistic avatar for user interaction. Macy successfully provided medication guidance on key symptoms, dosage, and precautions in less than 30 minutes at an affordable cost \cite{MacyAI}.\\
\indent\textbf{Treatment Suggestion.} LLM-based chatbots can also assist with treatment suggestions. Research in \cite{chen2023use} evaluates ChatGPT's alignment with National Comprehensive Cancer Network (NCCN) guidelines for breast, prostate, and lung cancer treatments. The authors develop four zero-shot prompt templates for 26 cancer diagnoses, generating 104 prompts without examples of correct responses. Three out of four board-certified oncologists assess ChatGPT's compliance with NCCN guidelines using five distinct criteria, yielding a total score of 520. The oncologists agree on 61.9\% (322 out of 520) of the scores. These results indicate that approximately two-thirds of the ChatGPT's treatment recommendations aligned with the established NCCN guidelines, highlighting its potential effectiveness in medical guidance. Another study \cite{koga2023evaluating} evaluates ChatGPT-3.5, ChatGPT-4, and Google BARD in predicting neuropathologic diagnoses using summaries from 25 neurodegenerative disorder cases. All these summaries are from Mayo Clinic Brain Bank Clinico-Pathological Conferences. The chatbots provide multiple diagnoses and rationales, which are later compared to the actual diagnoses. ChatGPT-3.5, ChatGPT-4, and Google BARD obtain primary diagnosis accuracies of 32\%, 52\%, and 40\%, respectively, while providing correct diagnoses in 76\%, 84\%, and 76\% of cases. This emphasizes the potential of LLM-based chatbots in neuropathology. Another similar study \cite{dhanvijay2023performance} evaluates the proficiency of ChatGPT 3.5, Google BARD (Experiment), and Bing Chat (Precise) in answering physiology case vignettes. Two physiologists prepare a total of 77 case vignettes, and two other experts verify them. Next, the two physiologists rate the chatbot's responses on a 0–4 scale, reflecting learning outcomes from basic to advanced understanding. ChatGPT scored highest at 3.19±0.3, followed by BARD at 2.91±0.5 and Bing at 2.15±0.6, indicating ChatGPT's superior performance in this context. In addition, one more article \cite{lim2023benchmarking} evaluates ChatGPT-3.5, ChatGPT-4, and Google BARD in responding to myopia-related queries. The study involves 31 myopia-related questions, which are classified into six domains: diagnosis, clinical presentation, pathogenesis, risk factors, treatment and prevention, and prognosis. Three pediatric ophthalmologists rated each chatbot's responses on a three-point scale (good, borderline, poor), with a majority consensus for the final scores. ‘Good’ responses undergo further evaluation for depth on a five-point scale, while ‘poor’ responses are prompted for self-improvement and reassessed for accuracy. ChatGPT-4 shows higher accuracy, with 80.6\% of its responses rated ‘good,’ compared to 61.3\% of ChatGPT-3.5 and 54.8\% of Google BARD. In terms of comprehensiveness, all three chatbots demonstrate high mean scores, with Google BARD scoring 4.35, followed by ChatGPT-4 at 4.23 and ChatGPT-3.5 at 4.11, on a scale of 5. They also exhibit notable self-correction abilities. 66.7\% of ChatGPT-4, 40\% of ChatGPT-3.5, and 60\% of Google BARD responses enhance post-correction. Besides, all the chatbots perform consistently in every domain, with ChatGPT-4 excelling in ‘treatment and prevention,’ obtaining 70\% ‘good’ ratings, notably higher than ChatGPT-3.5's 40\% and Google BARD's 45\%.\\
\begin{table*}[!ht]
\centering
\captionsetup{labelsep=newline,justification=centering}
\caption{LLM-based Chatbots in Various Sectors}
\label{tab:Table3}
\begin{tblr}{
  width = \textwidth,
  colspec = {|Q[c,m,0.085\linewidth]|Q[c,m,0.085\linewidth]|Q[c,m,0.397\linewidth]|Q[c,m,0.0655\linewidth]|Q[c,m,0.0655\linewidth]|Q[c,m,0.0655\linewidth]|Q[c,m,0.0655\linewidth]|},
  row{1} = {font=\bfseries, c},
  row{1-31} = {font=\linespread{1.1}\selectfont}, % Adjust line spread for all cells, if needed
  hlines,
  cell{2}{1} = {r=7}{halign=c,valign=m},
  cell{9}{1} = {r=7}{halign=c,valign=m},
  cell{16}{1} = {r=8}{halign=c,valign=m},
  cell{24}{1} = {r=4}{halign=c,valign=m},
  cell{28}{1} = {r=4}{halign=c,valign=m},
}
Sector     & Referenced Article   & Main Objective & ChatGPT    & BARD   & Bing Chat    & Other Chatbot(s)\\
Education  & \cite{vasconcelos2023enhancing}                 & Transforming STEM education & \CheckmarkBold & \textbf{-}       & \CheckmarkBold   & \textbf{-}\\
           & \cite{dao2023large}                 & Meeting the educational needs of Vietnamese students & \CheckmarkBold & \CheckmarkBold & \CheckmarkBold & \textbf{-}\\
                      & \cite{zhang2023one}      & Enhancing the academic writing skills of international scholars & \CheckmarkBold & \textbf{-}  & \textbf{-}  & \textbf{-}\\
                      & \cite{ChatGPTvsGrammarly}      & Assisting in developing a unique and engaging writing style & \CheckmarkBold & \textbf{-}  & \textbf{-}  & \textbf{-}\\
                      & \cite{tlili2023if}      & Aiding in curriculum design & \CheckmarkBold & \textbf{-}  & \textbf{-}  & \textbf{-}\\
                      & \cite{lo2023impact}      & Assisting with teaching preparation and evaluation tasks & \CheckmarkBold & \textbf{-}  & \textbf{-}  & \textbf{-}\\
                      & \cite{megahed2023generative}      & Drafting a syllabus for an undergraduate Statistics course & \CheckmarkBold & \textbf{-}  & \textbf{-}  & \textbf{-}\\
Research   & \cite{chandha2023setting}                 & Finding relevant literature & \CheckmarkBold & \textbf{-}  & \textbf{-}  & \textbf{-}\\
                      & \cite{aydin2022openai}      & Paraphrasing abstracts & \CheckmarkBold & \textbf{-}  & \textbf{-}  & \textbf{-}\\
                      & \cite{aydin2023google}       & Generating literature reviews & \textbf{-}  & \CheckmarkBold & \textbf{-}  & \textbf{-}\\
                      & \cite{macdonald2023can}       & Handling simulated datasets & \CheckmarkBold & \textbf{-}  & \textbf{-}  & \textbf{-}\\
                      & \cite{ChatGPTData}            & Assisting with data analysis tasks & \CheckmarkBold & \textbf{-}  & \textbf{-}  & \textbf{-}\\
                      & \cite{girotra2023ideas}       & Assisting with new idea generation & \CheckmarkBold & \textbf{-}  & \textbf{-}  & \textbf{-}\\
                      & \cite{temsah2023reflection}   & Analyzing the effects of COVID-19 from various dimensions  & \CheckmarkBold & \textbf{-}  & \textbf{-}  & \textbf{-}\\
Healthcare & \cite{gilson2023does}                 & Answering USMLE questions    & \CheckmarkBold & \textbf{-}  & \textbf{-}  & \CheckmarkBold \\
                      & \cite{hamidi2023evaluation}      & Answering clinical questions & \CheckmarkBold & \textbf{-}  & \textbf{-}  & \CheckmarkBold\\
                      & \cite{azizi2023evaluating}    & Answering questions on Atrial Fibrillation & \CheckmarkBold & \textbf{-}  & \CheckmarkBold & \textbf{-}\\
                      & \cite{MacyAI}      & Pioneering intelligent robotic assistance & \CheckmarkBold & \textbf{-}  & \textbf{-}  & \textbf{-}\\
                      & \cite{chen2023use}      & Giving suggestions for cancer treatments & \CheckmarkBold & \textbf{-}  & \textbf{-}  & \textbf{-}\\
                      & \cite{koga2023evaluating}      & Predicting neuropathologic diagnoses & \CheckmarkBold & \CheckmarkBold & \textbf{-}  & \textbf{-}\\
                      & \cite{dhanvijay2023performance}      & Answering physiology case vignettes & \CheckmarkBold & \CheckmarkBold & \CheckmarkBold & \textbf{-}\\
                      & \cite{lim2023benchmarking}      & Answering myopia-related queries & \CheckmarkBold & \CheckmarkBold & \textbf{-}  & \textbf{-}\\
Miscellaneous Applications (Software Engineering) & \cite{meyer2023chatgpt}                 & Providing programming support    & \CheckmarkBold & \textbf{-}  & \textbf{-}  & \textbf{-}\\
                      & \cite{surameery2023use}    & Fixing bugs and providing clear explanations & \CheckmarkBold & \textbf{-} & \textbf{-}  & \textbf{-}\\
                      & \cite{belzner2023large}      & Performing software engineering tasks & \CheckmarkBold & \CheckmarkBold & \textbf{-}  & \textbf{-}\\
                      & \cite{nejjar2023llms}      & Generating Java code for matrix multiplication & \CheckmarkBold & \CheckmarkBold & \CheckmarkBold & \CheckmarkBold\\
Miscellaneous Applications (Finance) & \cite{dowling2023chatgpt}                 & Providing insights into financial research    & \CheckmarkBold & \textbf{-}  & \textbf{-}  & \textbf{-}\\
                      & \cite{gill2023chatgpt}      & Analyzing economic data and giving investment suggestions & \CheckmarkBold & \textbf{-}  & \textbf{-}  & \textbf{-}\\
                      & \cite{lakkaraju2023can}      & Providing investment advice & \CheckmarkBold & \CheckmarkBold & \textbf{-}  & \textbf{-}\\
                      & \cite{altan7science}      & Supporting analysts in strategic decision-making & \textbf{-}  & \textbf{-}  & \CheckmarkBold & \textbf{-}                      
\end{tblr}
\vspace{-2mm}
\end{table*}
\indent Overall, the effectiveness of LLM-based chatbots like ChatGPT, BARD, Bing Chat, and Claude in the medical field is evident. Their diverse applications, from answering complex medical questions to providing personalized patient education and treatment suggestions, underscore the vital role they can play in improving the healthcare domain and patient interaction. Additionally, advancements like GPT-4's ability to handle multilevel prompts, images, and documents suggest these chatbots will soon be able to analyze multimedia, which will further elevate the medical sector.
\vspace{-2mm}
\subsection{Miscellaneous Applications}
Moving beyond their roles in education, research, and healthcare, LLM-based chatbots are also gaining popularity in sectors like software engineering and finance. Here, we discuss how LLM-based chatbots offer more effective and scalable solutions across these industries with unparalleled efficiency and customization.\\
\indent\textbf{Software Engineering.} In contrast to traditional command-based software development support, interacting with LLM-based chatbots focuses on the user’s intent and fosters a conversational approach \cite{belzner2023large}. Developers discuss their needs or the outcomes they seek without specifying the methods to achieve them. This transition facilitates LLM-based chatbots for numerous tasks, such as writing code, finding and fixing errors, and testing software quality. For example, a study in \cite{meyer2023chatgpt} explores how ChatGPT serves as an interactive teaching tool, providing language selection advice, code syntax guidance, insights on best practices, library or package recommendations, alternative method suggestions, IDE introductions, and programming environment advice. Besides, ChatGPT can fix bugs and provide clear explanations of complex topics, ensuring a comprehensive learning experience \cite{surameery2023use}. Unlike searching through Google or sites like Stack Overflow and GeeksforGeeks for coding guidance, ChatGPT provides learners with straightforward and often practical solutions to their programming questions. Another article \cite{belzner2023large} conducts a case study on a “Search and Rescue” scenario to demonstrate the application of BARD and ChatGPT in software engineering tasks. BARD is noted for offering abstract, high-level advice that emphasizes overarching concepts and strategies, such as proposing theoretical tests and discussing complex algorithms. On the other hand, ChatGPT provides detailed, feasible solutions, focusing on specific coding practices and unit testing with practical frameworks. This distinction highlights BARD’s strength in strategic guidance and ChatGPT’s proficiency in delivering precise, implementation-ready solutions, underscoring their complementary roles in software development tasks. \cite{nejjar2023llms} further evaluates the ability of several LLM-based chatbots to generate Java code for matrix multiplication, with a particular focus on multi-threading. The test includes ChatGPT 3.5 and 4, BARD, Bing Chat, YouChat, GitHub Copilot, and GitLab Duo. Most of these chatbots generate the correct code on the first attempt, except for Google BARD, which needs human assistance. YouChat stands out with the fastest code (446 ms), while Bing Chat performs the slowest (1899 ms). However, it is worth noting that GitHub Copilot, Bing Chat, and YouChat tend to create brief but undetailed code.\\
\indent In addition, a user can ask these chatbots to explain a piece of code. The chatbots will explain each part, including variables and commands. They can also summarize what the code does, improving the clarity and understanding of the code. To conclude, LLM-based chatbots offer a transformative approach to software engineering by enabling intent-based and conversational interactions that cover a range of tasks from code generation to debugging, software testing, and providing educational support. This capability not only increases productivity but also makes software engineering expertise more accessible to programmers at every level.\\
\begin{table*}[!ht]
\centering
\captionsetup{labelsep=newline,justification=centering}
\caption{Chatbot Challenges: Knowledge and Data Viewpoints}
\label{tab:Table4}
\begin{tblr}{
  width = \textwidth,
  colspec = {|Q[c,m,0.075\linewidth]|Q[c,m,0.22\linewidth]|Q[c,m,0.375\linewidth]|Q[c,m,0.1035\linewidth]|Q[c,m,0.1035\linewidth]|},
  row{1} = {font=\bfseries, c},
  row{1-11} = {font=\linespread{1.1}\selectfont}, % Adjust line spread for all cells, if needed
  hlines,
  cell{1}{4} = {c=2}{}, % This combines the fourth and fifth cell in the first row for "Viewpoint"
  cell{2}{4} = {c=1}{halign=c,valign=m}, % These two cells specify the individual columns
  cell{2}{5} = {c=1}{halign=c,valign=m}, % under "Viewpoint" for "Knowledge" and "Data"
  cell{1}{1} = {r=2}{halign=c,valign=m},
  cell{1}{2} = {r=2}{halign=c,valign=m},
  cell{1}{3} = {r=2}{halign=c,valign=m},
  cell{3}{1} = {r=3}{halign=c,valign=m},
  cell{6}{1} = {r=4}{halign=c,valign=m},
  cell{10}{1} ={r=3}{halign=c,valign=m},
}
Perspective                & Challenge                       & Discussion Topic & \SetCell[c=2]{c} Viewpoint \\
                           &                                 &                  & Knowledge & Data \\
From a Technical Perspective  & Knowledge Recency     & Challenges in maintaining up-to-date knowledge    & \textbf{-}  & \CheckmarkBold   \\
                              & Logical Reasoning     & Performance gap in multi-step reasoning questions & \CheckmarkBold & \textbf{-}    \\
                              & Hallucination         & Generation of incorrect and unreliable responses  & \CheckmarkBold  &    \textbf{-} \\
From an Ethical Perspective   & Transparency          & Lack of clarity in chatbot reasoning processes    & \CheckmarkBold  & \CheckmarkBold   \\
                              & Bias                  & Data bias in chatbot training and responses       & \CheckmarkBold  & \CheckmarkBold   \\
                              & Privacy Risks         & Privacy concerns and data protection issues       & \CheckmarkBold  & \CheckmarkBold   \\
                              & Unfairness            & Linguistic and economic unfairness in chatbot accessibility & \textbf{-}  & \CheckmarkBold   \\
From a Misuse Perspective     & Academic Misuse       & Challenges of preserving academic integrity       & \CheckmarkBold & \textbf{-}    \\
                              & Over-reliance         & Impact on the skills of critical thinking                           & \CheckmarkBold & \textbf{-}    \\
                              & Distribution of Wrong Information         & Potential spreading of misleading information       & \CheckmarkBold & \textbf{-}
\end{tblr}
\vspace{-0.15mm}
\end{table*}
%\vspace{-2mm}
\indent\textbf{Finance.} LLM-based chatbots are making breakthroughs in the finance sector. Their ability to match resources with customer needs enhances service effectiveness, and they assist employees in managing their daily workload more efficiently. For example, a study in \cite{dowling2023chatgpt} investigates the applications of ChatGPT in the finance industry. First, it looks at the potential of using machine learning to analyze financial data and its applications in finance. Next, it presents the “Bananarama Conjecture,” which suggests that ChatGPT can provide better insights into financial research than conventional methods. Another study \cite{gill2023chatgpt} explores how ChatGPT effectively analyzes financial information to identify trends, market opinions, and movements. Its ability to analyze economic data and provide investment recommendations is a boon to companies and financiers. \cite{lakkaraju2023can} further evaluates ChatGPT and BARD in the finance domain for providing investment advice in different languages and dialects, including English, African American Vernacular English (AAVE), and Telugu. Compared to ChatGPT, BARD offers varied responses through multiple drafts but fails to refine content with each query. Besides, BARD does not understand Telugu, showing multi-lingual limitations. ChatGPT, on the other hand, consistently corrects errors and adapts to AAVE over time, though it struggles with Telugu language accuracy. The study reveals BARD's lower personalization rate (53\%) and higher error rates (15.38\%) compared to ChatGPT (46.15\% and 7.69\%, respectively), with ChatGPT also facing a 15.38\% rate of grammatical errors. Despite these issues, their potential for analyzing vast financial data is evident, showcasing significant capabilities in complex information handling. One more article \cite{altan7science} evaluates Bing Chat's role in assisting analysts with investment advice and portfolio recommendations. Bing Chat analyzes financial documents from 2019 to 2022 to recommend a stock portfolio from the BIST100, selecting six specific companies. It also guides portfolio composition, suggesting a specific number of stocks based on portfolio size. Overall, Bing Chat offers valuable financial insights and recommendations, supporting analysts in strategic decision-making.\\
\indent To conclude, these are just a few of the many applications LLM-based chatbots find across different fields. As technology advances, these chatbots are poised to become deeply embedded in our lives, reshaping our interactions with technology and each other. Also, the growing use of chatbots, spurred by these advancements in AI, is a response to evolving consumer preferences and the need for improved interactive technologies. Adding to our discussion, Table \ref{tab:Table3} provides an overview of specific chatbots used in different sectors for various purposes. It highlights their roles and links them to their respective referenced articles. Readers interested in particular chatbot implementations can refer to these articles for more information.
\vspace{-2mm}
\section{Open Challenges}
\label{sec:Challenges}
As LLM-based chatbots evolve, they encounter numerous challenges across different domains. In this section, we discuss these challenges, providing an insightful overview from technical, ethical, and misuse perspectives. Table \ref{tab:Table4} categorizes the challenges from each perspective according to their viewpoints on knowledge or data, offering a structured outline that clarifies the context for the readers.
\vspace{-2mm}
\subsection{From a Technical Perspective}
Here, we explore the technical limitations of LLM-based chatbots with a specific focus on knowledge recency, logical reasoning, and hallucination.\\
\indent\textbf{Knowledge Recency.} Maintaining up-to-date knowledge presents a notable challenge for the LLM-based chatbots, as they often struggle with tasks that require information beyond their most recent training. While updating LLMs regularly with new data is a straightforward solution, it is expensive and poses the risk of catastrophic forgetting during incremental training. This makes adjusting the built-in knowledge of LLMs a complex task \cite{dai2021knowledge, meng2022locating}. Besides, the lack of diverse, high-quality data sources presents future limitations \cite{bender2021dangers, villalobos2022will}.\\
\indent\textbf{Logical Reasoning.} Chatbots lack rational human thinking. As a result, they can neither think nor reason like human beings \cite{saghafian2023analytics, agarwal2023analysing}. A study in \cite{cai2023performance} evaluates the performance of ChatGPT-3.5, ChatGPT-4, and Bing Chat using 250 questions from the Basic Science and Clinical Science Self-Assessment Program and then compares them to those of human participants. Humans achieve an average accuracy of 72.2\%. ChatGPT-3.5 scores the lowest at 58.8\%, while ChatGPT-4 and Bing Chat show similar results, scoring 71.6\% and 71.2\%, respectively. In single-step reasoning questions, all three chatbots perform well, with ChatGPT-3.5, ChatGPT-4, and Bing Chat achieving accuracies of 68.4\%, 80.0\%, and 81.0\%, respectively. However, their performances significantly decline in multi-step reasoning questions, where ChatGPT-3.5 scores only 40.0\%, while ChatGPT-4 and Bing Chat score 64.5\% and 60.0\%, respectively. Another recent paper \cite{sutcliffe2023chat} evaluates the logical reasoning skills of BARD. The authors use the TPTP problem PUZ001+1 to pose a specific question and utilize tools from the TPTP World to analyze BARD's responses. The findings reveal that BARD's reasoning is flawed, leading to incorrect conclusions from the provided data, which is attributed to the lack of formal reasoning abilities. However, the study also acknowledges that this test focuses on a particular reasoning task, and BARD might show better results in other tasks. \cite{nguyen2023evaluation} further highlights the limited logical reasoning abilities in mathematics, as BARD performs poorly on the Vietnamese National High School Graduation Examination (VNHSGE) mathematics test, showing a mere 38.8\% accuracy.\\
\indent\textbf{Hallucination.} LLM-based chatbots face a notable challenge in producing factual texts due to hallucination \cite{zhao2023survey, bang2023multitask}, where the generated information either contradicts existing sources (intrinsic hallucination) or cannot be confirmed by available sources (extrinsic hallucination). In simpler terms, a hallucination is a confident response by the chatbot that is neither correct nor reliable. For example, \cite{meyer2023chatgpt} highlights how ChatGPT creates entirely fictional publications when asked to find relevant citations for a review paper. Another study \cite{rudolph2023war} compares the performance of ChatGPT-3.5, GPT-4, Bing Chat, and BARD by asking them about the most-cited ChatGPT articles in higher education. The results are disappointing across all chatbots. For instance, ChatGPT gives five entirely irrelevant references dating back to 1975. GPT-4 shows a marginal improvement, but Bing Chat and BARD offer entirely fictional references. Concerns with chatbot use in healthcare also include hallucinations, where outputs sound convincingly plausible but are factually inaccurate. The study in \cite{chen2023use} evaluates the recommendations of ChatGPT for breast, prostate, and lung cancers according to the 2021 NCCN guidelines. The authors prepare four zero-shot prompt templates to generate 104 prompts from 26 cancer diagnoses without examples of correct responses. Then, three board-certified oncologists evaluate the responses to these 104 prompts using five criteria for a total of 520 scores. Among these, 13 of the 104 outputs (12.5\%) are identified as hallucinations, meaning they do not align with any recommended treatments. Additionally, another study \cite{azizi2023evaluating} demonstrates that while ChatGPT and Bing Chat provide accurate answers to queries about Atrial Fibrillation, some responses include fictitious or incorrectly cited references. Bing Chat, in comparison to ChatGPT, exhibits higher accuracy in responses but a comparable frequency of inaccurate references. Although GPT-4 shows improvements in reducing hallucinations compared to the previous versions, there is still a need for continued research to minimize this issue further.\\
\indent Apart from these, LLM-based chatbots also have a consistency issue, often generating different responses for the same input \cite{zhang2023one}. Researchers are trying to improve this through prompt engineering \cite{wang2023can}. It is also important to note that these chatbots lack self-awareness, emotions, or subjective experiences, even though they can answer questions and generate coherent text \cite{borji2023categorical}. There is an ongoing debate about whether machines can truly have self-awareness, with no clear definition or measurement methods established yet.
\begin{comment}
    \subsection{Figures}
Fig. 1 is an example of a floating figure using the graphicx package.
 Note that $\backslash${\tt{label}} must occur AFTER (or within) $\backslash${\tt{caption}}.
 For figures, $\backslash${\tt{caption}} should occur after the $\backslash${\tt{includegraphics}}.
\end{comment}
\vspace{-3.05mm}
\subsection{From an Ethical Perspective}
In this subsection, we discuss the ethical issues of LLM-based chatbots, highlighting key areas such as transparency, bias, privacy risks, and unfairness.\\
\indent\textbf{Transparency.} LLMs are often described as black box models due to the complexity of their answer-generation processes from input queries. As a result, LLM-based chatbots exhibit a lack of transparency, making it difficult to understand the reasoning behind a specific output or decision \cite{rudin2019stop}. For example, in the healthcare domain, this transparency issue of LLM-based chatbots is a significant concern since health responses and genetic factors vary greatly across populations \cite{rao2023evaluating}. Moreover, the transparency of training data, which may not be verified for domain-specific accuracy, leads to a ‘garbage in, garbage out’ issue. This is true for models like GPT-3.5, which does not verify the accuracy of its training data \cite{brown2020language}. Additionally, OpenAI's transition from a non-profit to a business-centric organization has raised concerns about its transparency in disclosing the technical details of its advancements. For instance, the GPT-4 technical report \cite{openai2023gpt4} mainly focuses on its improved performance over previous models but falls short of explaining the underlying technical methods used to achieve these enhancements.\\
\indent\textbf{Bias.} Another concern regarding the LLM-based chatbots is bias. This occurs when the models are trained on biased data, which may represent racial, gender, or socioeconomic inequalities in society. As noted in \cite{schramowski2022large}, large pre-trained models mimicking natural languages can repeat these biases. Moreover, chatbots' responses are shaped by the input they receive. If users frequently ask biased questions, the model may learn and replicate them \cite{fraiwan2023review}. Also, when models are fine-tuned to optimize specific metrics like accuracy or user engagement, there is a risk of algorithmic bias, where responses may align with these goals, regardless of inherent biases. In medical treatment, using chatbots trained on biased data can lead to inaccurate medical results, potentially causing harm to both patients and communities. For instance, a chatbot might misdiagnose a medical condition due to biased training and recommend the wrong treatments. A study in \cite{phillips2019assessment} highlights an AI system designed for skin disease diagnosis that generates a high rate of false positives. This issue leads to unnecessary biopsy procedures and increases anxiety among patients. Another study \cite{oca2023bias} observes a bias in Bing Chat and Google BARD, with tendencies to recommend mostly male ophthalmologists. Besides, in academic research, these chatbots might yield inaccurate or biased outcomes. For example, a chatbot trained with biased data may generate incorrect findings in social science research, resulting in erroneous conclusions that could negatively impact marginalized groups \cite{kooli2023chatbots}. Furthermore, \cite{mcgee2023chat} finds that ChatGPT shows political bias, favoring liberal over conservative views when creating an Irish limerick. In addition, ChatGPT is also found to have a left-wing liberal bias in reviewing political elections in democratic countries \cite{hartmann2023political}.\\
\begin{comment}
    \section{Tables}
Note that, for IEEE-style tables, the
 $\backslash${\tt{caption}} command should come BEFORE the table. Table captions use title case. Articles (a, an, the), coordinating conjunctions (and, but, for, or, nor), and most short prepositions are lowercase unless they are the first or last word. Table text will default to $\backslash${\tt{footnotesize}} as
 the IEEE normally uses this smaller font for tables.
 The $\backslash${\tt{label}} must come after $\backslash${\tt{caption}} as always.
 
\begin{table}[!t]
\caption{An Example of a Table\label{tab:table1}}
\centering
\begin{tabular}{|c||c|}
\hline
One & Two\\
\hline
Three & Four\\
\hline
\end{tabular}
\end{table}
\end{comment}
\indent\textbf{Privacy Risks.} Moving from transparency and bias, another paramount issue of the LLM-based chatbots is user privacy and data protection. Italy has recently imposed a ban on ChatGPT after a data breach, highlighting privacy issues and the absence of age verification, which pose a risk of exposing minors to inappropriate content \cite{mccallum2023chatgpt}. These chatbots are trained on large datasets that often include sensitive user information, such as chat logs and personal details, which can lead to privacy concerns. Moreover, chatbots can produce personalized output depending on user queries. For instance, if users input confidential information like health or financial data, the chatbot might accidentally reveal that information to others \cite{fraiwan2023review}. Furthermore, it is crucial to acknowledge that OpenAI may collect any personal information included in the input according to their privacy policy \cite{OpenAIPP}. Therefore, such misuse of personal information can lead to harmful consequences for users, particularly when it falls into the hands of criminals.\\
\indent\textbf{Unfairness.} Biases in training data can cause language models to increase unfairness in learning, often marginalizing smaller groups. Since most research on large language models primarily serves English speakers, a significant research gap persists for other languages, introducing a layer of linguistic unfairness. This could potentially smooth the educational path for English-speaking users while sidelining those who speak other languages by restricting their access to these technological advances \cite{kasneci2023chatgpt}. Furthermore, the economic unfairness arising from the cost of accessing LLM-based chatbots, like ChatGPT Plus, could potentially widen the educational gap in unprecedented ways \cite{kasneci2023chatgpt, zhang2023one, liebrenz2023generating}.\\
\indent Therefore, prioritizing ethical standards in creating and using LLM-based chatbots is essential. Addressing transparency, bias, privacy risks, and unfairness is necessary to maintain ethical integrity while ensuring user trust and safety.
\vspace{-2mm}
\subsection{From a Misuse Perspective}
Here, we address the practical challenges of LLM-based chatbots, focusing on areas like academic misuse, over-reliance, and the distribution of wrong information, highlighting their impact in real-world scenarios.\\
\indent\textbf{Academic Misuse.} LLM-based chatbots are commonly misused in academic writing, where students and researchers might use their generated content in exams and research papers without proper citation. Many institutions have banned these tools, citing concerns that they compromise evaluation standards and the value of education \cite{zhang2023one, shiri2023chatgpt}. A study in \cite{king2023conversation} investigates the growing concern about plagiarism in higher education and the use of ChatGPT for cheating. Besides, \cite{motlagh2023impact, rudolph2023war, khalil2023will} show that ChatGPT can generate complex and authentic content, often not detected by standard anti-plagiarism software like iThenticate or Turnitin \cite{susnjak2022chatgpt, wiggers2023openai, gimpel2023unlocking}, which further questions the probity of online exams.\\
\indent\textbf{Over-reliance.} Another rising concern is the increasing dependency on LLM-based chatbots in research. For example, \cite{aydin2023google} demonstrates the use of Google BARD in generating literature reviews. Although the texts produced by BARD initially contained some plagiarism, it could be resolved using a paraphrasing tool. Another study \cite{dao2023chatgpt} evaluates the performance of Bing Chat in catering to the academic needs of Vietnamese students in a range of subjects, including Mathematics, English, Physics, Chemistry, Biology, Literature, History, Geography, and Civic Education. The results show that Bing Chat outshines ChatGPT in most subjects, except in literature, where ChatGPT performs better. Moreover, Bing Chat's accessibility in Vietnam and its ability to include hyperlinks and citations in the responses further emphasize its advantage. With the rise of these intelligent systems, students get access to substantial computing power that aids them largely in their academic work \cite{shiri2023chatgpt}. However, this dependency and empowerment often come at the cost of weakening the ability to think critically and independently.\\
\indent\textbf{Distribution of Wrong Information.} As mentioned earlier, LLM-based chatbots operate as a black box, making it difficult to interpret how they process and make decisions \cite{tan2023generative, abramoff2022foundational}. Unless specifically asked, responses are not referenced or explained, and the reliability of any explanation is uncertain. Consequently, a significant issue with these LLM-based chatbots is their potential to spread wrong information as if it were true \cite{zhang2023one, zhuo2023exploring, borji2023categorical}. For example, a study in \cite{thirunavukarasu2023trialling} highlights that users might misuse ChatGPT to falsely suggest medical diagnoses by offering information that seems to be accurate and reliable. Another survey \cite{zhang2023one} shows how these chatbots can be used to create numerous false articles for blogs, media, newspapers, or the internet. These articles may look authentic, but it is possible that they are fake and do not exist at all, which makes it more difficult to distinguish fact from fiction. Moreover, Microsoft’s integration of Bing Chat into its search engine could significantly speed up the propagation of false information on the internet. Without proper controls, this rapid circulation of misinformation could be detrimental to public information security. Besides, several recent articles \cite{sebastian2023privacy, al2023chatgpt, derner2023beyond} explore the potential vulnerabilities and threats associated with these chatbots, including various attack vectors, information extraction, and the creation of harmful content. \cite{zhang2023one} highlights that hackers might exploit ChatGPT's programming skills to develop malicious software \cite{dash2023chatgpt}, like viruses or Trojans, for network assaults, data theft, or to hijack other computer systems, which can cause significant harm to users. In addition, trolls could manipulate ChatGPT with targeted prompts and produce harmful content to attack other users \cite{zhuo2023exploring}.\\
\indent To conclude, while LLM-based chatbots offer valuable advantages in various domains, their potential for misuse poses substantial risks. Users must ensure proper citation to prevent academic misconduct, avoid over-reliance on these tools to maintain critical and independent thinking skills, and carefully validate the accuracy of the information provided.
\vspace{-2mm}
%Privacy: [8. CS - 7]
\section{Future Outlook}
\label{sec:Outlook}
In this section, we explore the future outlook of LLM-based chatbots, from technical improvements for efficiency and sustainability to ethical considerations that guide their responsible usage.
\begin{figure}[t]
\centerline{\includegraphics[width=\linewidth]{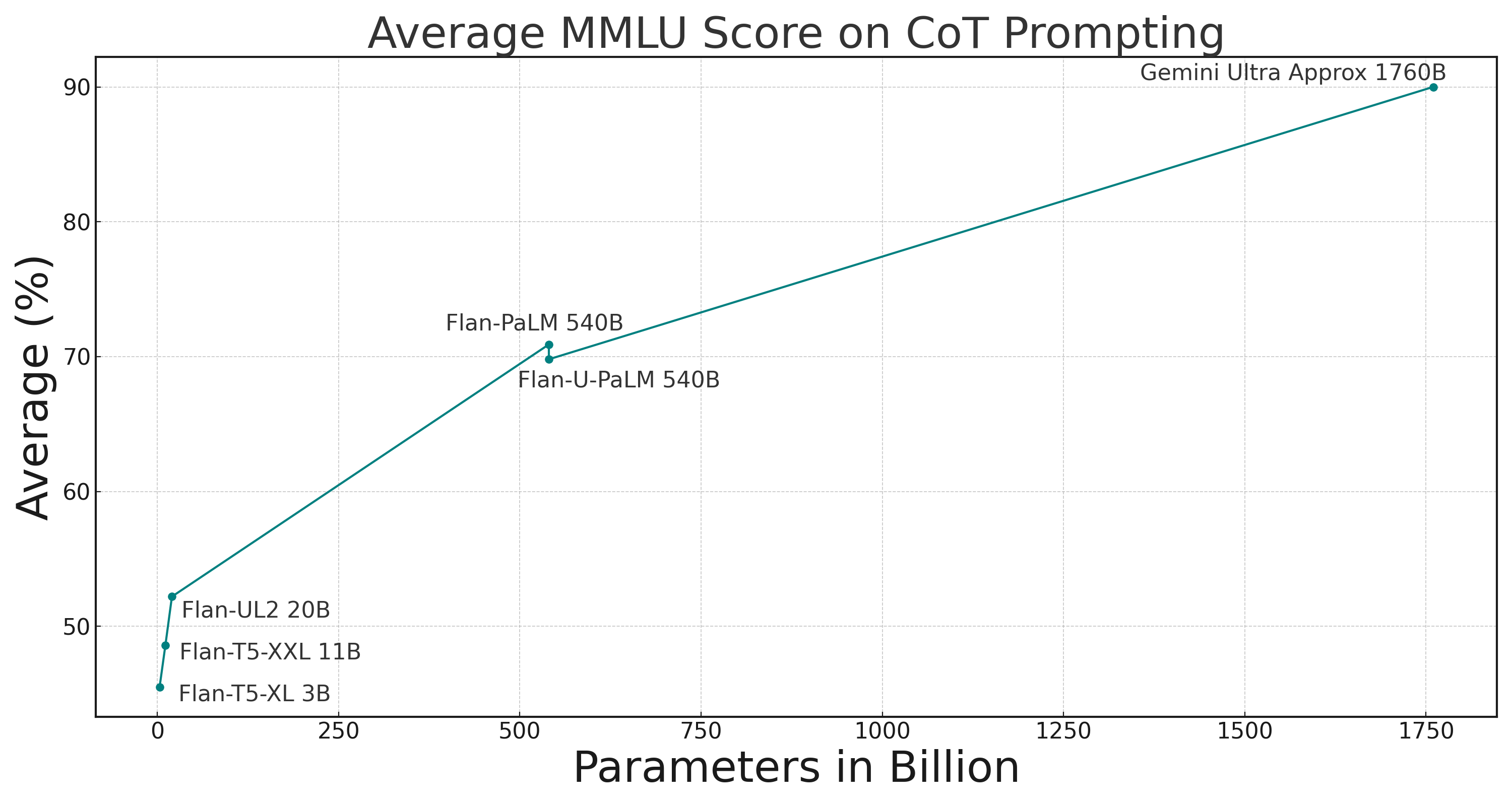}}
\caption{Correlation between model size and average MMLU scores \cite{MMLUSOTA2}.}
\label{fig:MMLU1}
\vspace{-6mm}
\end{figure}
\vspace{-2mm}
\subsection{Technical Improvements}
Here, we focus on model compression and optimization for enhanced efficiency, the use of green AI technologies in addressing environmental issues, advancements in prompt engineering, and the emergence of multimodal capabilities extending beyond text.\\
%In this section, we focus on model compression and optimization for enhanced efficiency, the use of green AI technologies in addressing environmental issues, the future of prompt engineering for better interaction dynamics, and the emergence of multimodal capabilities extending beyond text.
\indent\textbf{Model Compression and Optimization.} Transformer-based language models demonstrate enhanced capabilities as their parameter count increases \cite{wei2022emergent, kaplan2020scaling}. Notable developments such as in-context learning \cite{brown2020language} and chain-of-thought (CoT) prompting \cite{wei2022chain} become increasingly evident when models surpass certain size thresholds. For example, Fig. \ref{fig:MMLU1} illustrates the average scores on CoT prompting for the Massive Multitask Language Understanding (MMLU) benchmark across models of various sizes \cite{MMLUSOTA2, hendrycks2020measuring}. The graph shows that, as the number of parameters increases from 3 billion in Flan-T5-XL to approximately 1760 billion in Gemini Ultra, there is a substantial rise in the MMLU scores. However, despite the advancements these LLMs bring to natural language processing, as guided by scaling laws \cite{kaplan2020scaling}, their massive size, often over 100 billion parameters, poses practical challenges. They include high costs for storage, distribution, and deployment. To address them, future research should focus more on model compression and optimization \cite{kaplan2020scaling}. These are vital for using modern chatbots in real-world scenarios. There are several approaches to this. Distillation-based methods train a smaller model using data from a larger one \cite{jiao2019tinybert, hinton2015distilling, sun2020mobilebert}. Pruning-based methods decrease the model size by removing redundant weights \cite{gordon2020compressing, chen2020lottery}, whereas quantization reduces the storage size of model weights \cite{shen2020q, bai2020binarybert}. However, these methods may require specialized hardware. Interested readers can refer to \cite{cheng2017survey} for more details.\\
\indent\textbf{Green AI.} The growing usage of LLM-based chatbots has brought attention to environmental issues, given their dependency on extensive computing resources for training. These chatbots, built upon large pre-trained models, often carry inherent biases from multiple data sources, making bias mitigation a challenging task due to the complex nature of their development. In contrast, traditional chatbots exhibit less bias but cannot generate fluent and diverse natural language content. Enhancing LLMs with knowledge graphs (KGs) might improve their knowledge base \cite{pan2023unifying, sun2023think}. However, this integration does not yet offer an entirely transparent reasoning process. That said, the increasing size and resource demands of machine learning models have brought green AI into focus \cite{kuo2023green, schwartz2020green}. Green learning (GL) technologies, aiming for eco-friendly AI systems with smaller, less complex models, could be crucial in developing chatbots with simpler reasoning processes and reduced resource requirements. They can also offer performance on par with Deep Learning (DL) in various applications \cite{kuo2023green}. A possible design for GL-based chatbots might involve splitting LLMs into two modules: one GL-based module dedicated to user interaction, managing tasks related to natural language understanding and generation, and another focused on knowledge storage, expansion, and reasoning through KGs. This approach might pave the way for more transparent, scalable, and unbiased chatbots, contributing to the development of fairer AI systems.\\
\indent\textbf{Prompt Engineering.} Prompt engineering is also becoming essential for the effective use of AI chatbots, impacting a wide range of applications, from everyday tasks to complex data analysis. The quality of prompts is crucial, as it determines the relevance and accuracy of the AI’s responses, embodying the idea that the output quality is only as good as the input. Effective prompts typically consist of four elements. They are context setting, specific instructions, format or structure, and optional examples. The context setting provides background information that helps AI understand the context of the responses. Specific instructions clarify the task or question, aiming to obtain relevant responses from the LLM. The formatting or structural feature determines the structure of the responses, including word count, bullet points, or visual elements like tables and graphs. Finally, optional examples, ranging from zero-shot to few-shot prompting, further improve the response quality. These examples illustrate the ideal format or structure of the responses. A recent study \cite{wei2022chain} shows that prompts can induce reasoning-like responses in the LLMs. Undoubtedly, this will lead to further innovations in prompting techniques. To conclude, the growing importance of prompt engineering, highlighted by an increase in recent publications, implies that achieving artificial general intelligence (AGI) might require more creative approaches than just increasing model size and data volume. Future work in this area is expected to explore new methods toward this goal.\\
\indent\textbf{Multimodality.} Integrating LLM-based chatbots with computer vision and robotics expands the capabilities of these systems beyond traditional text-based interactions. For instance, ChatGPT, Claude, and Bing Chat can generate descriptions of visual content from user inputs, address queries about images, and handle documents, including PDFs and CSVs. BARD, on the other hand, excels in visual content interpretation but lacks the functionality to process documents. Another area of exploration is the advancement of transfer learning techniques, enabling ChatGPT and other chatbots to assimilate knowledge from both linguistic and visual domains effectively. Pre-training the model on extensive datasets, such as the Conceptual Captions dataset \cite{sharma2018conceptual}, which combines text and image data, could deepen chatbots’ understanding of the relationships between language and visual information. The prospective integration of chatbots with computer vision technology heralds a new era of possibilities in AI. These include artistic creations like painting \cite{guo2023can}, intelligent vehicle operation \cite{du2023chat, gao2023chat, zhang2023hivegpt}, industrial automation \cite{wang2023chat}, and visually interactive conversational systems \cite{wang2023chatgpt}. Beyond computer vision, integrating these chatbots with chemical systems using technologies like SMILES \cite{weininger1988smiles} could revolutionize how chemical compositions are interpreted and interacted with. This integration could also simplify complex chemical analysis and enhance research capabilities in fields like pharmacology and materials science.\\
\indent In summary, the future of LLM-based chatbots lies in optimizing model efficiency, integrating green AI for environmental sustainability, enhancing prompt engineering for better interaction dynamics, and embracing multimodal capabilities that extend beyond mere textual communications.
\vspace{-2mm}
\subsection{Ethical Guidelines and Responsible Usage}
Here, we discuss the critical aspects of ethical considerations and responsible usage of LLM-based chatbots. We explore privacy and data protection for secure user interactions, highlight language diversity and equal rights to promote universal accessibility, and address academic and medical protocols to ensure fairness and responsibility in education and healthcare.\\
\indent\textbf{Privacy and Data Protection.} Training the LLM-based chatbots requires large datasets, often containing sensitive user information like chat logs and personal details. Thus, it is crucial to ensure the privacy and security of user data to retain trust in the technology \cite{koubaa2023exploring, harkous2016pribots, hasal2021chatbots}. Also, in healthcare, anonymizing and protecting patient data during training is mandatory to comply with privacy laws like HIPAA \cite{mesko2023imperative}. Researchers and developers must implement strict privacy and security measures, such as encryption, data anonymization, and controlled access to data. Besides, when using these technologies in healthcare, patient approval, transparency, and ethical standards are all equally important \cite{mesko2023imperative}.\\
\indent\textbf{Language Diversity and Equal Rights.} Despite advancements in AI tools like GPT-4 and ChatGPT, there is a significant performance gap in non-English languages due to limited datasets \cite{ahmed2022freely, ahmed2022arabic}. Consequently, developers of generative AI tools face the challenge of ensuring these technologies are inclusive, fair, and effective across diverse languages and user needs. Therefore, in the future, developers must focus on creating AI technologies that can serve a wide range of users, including those who are underprivileged or have disabilities, by providing multimodal interaction options. In addition, it is necessary to avoid bias and unfairness in the training data. Because chatbots like ChatGPT might unintentionally support stereotypes or discriminate against certain people if they use biased data. On the other hand, fairness means treating every user equally and not allowing their background to affect the service they receive. Therefore, the developers must continue monitoring the chatbots even after the training phase is complete, ensuring any biases are promptly identified and fixed. It will guarantee equal access to information and services for all users, maintaining ethical standards and fairness throughout the development and use of chatbots \cite{koubaa2023exploring, beattie2022measuring, dwivedi2023so}. Future works should also consider user-focused design principles, emphasizing social, emotional, cognitive, and pedagogical aspects \cite{kuhail2023interacting}. Drawing inspiration from platforms like Duolingo and Khan Academy, developers can harness ChatGPT and other chatbots to offer personalized learning experiences and real-time feedback at various levels of education. This includes using chatbots for interactive clinical communication modules and peer-to-peer learning experiences, enhancing the depth and practicality of specialized training.\\
\indent \textbf{Academic and Medical Protocols.} One rising concern is the misuse of LLM-based chatbots in education. Using these chatbots in education requires careful balance because, while they offer valuable insights, they cannot replace the unique human abilities to create and think critically. As noted in \cite{hosseini2023ethics, sallam2023chatgpt, van2023chatgpt}, banning these chatbots is not a feasible solution. Instead, there should be rules and regulations for accountability, integrity, transparency, and honesty. Several studies \cite{evans2023chatgpt, lund2023chatgpt, liebrenz2023generating} investigate the use of ChatGPT in academic writing, highlighting concerns about authorship, transparency, and bias that demand the establishment of ethical guidelines and commitment to best practices. Careful consideration of which academic skills are crucial for researchers is needed. The academic community should initiate the development and responsible use of LLM-based chatbots in research, guided by a comprehensive code of ethics, to ensure that moral and professional standards are maintained. Besides, integrating critical thinking and problem-solving exercises into the curriculum can effectively guide students in developing creative skills from an early stage \cite{zhang2023one}.\\
\indent In this broader context of AI integration in education, medical institutions, despite the widespread adoption of chatbots at various levels, from primary schools to universities, are still in the early stages of utilizing this technology. With the increasing role of generative AI tools and LLM-based chatbots in education, educators and students in the medical field face unique challenges and opportunities. Administrators must develop strategies to incorporate new technologies into medical education responsibly. These strategies include creating guidelines for AI tool usage in assignments, using text detection tools like Originality AI, Turnitin, and ZeroGPT, and conducting training for effective and ethical use of AI \cite{rudolph2023chatgpt}. Educators, on the other hand, should embrace these technologies and integrate them into the medical curriculum. This involves updating course content to cover AI’s role in medicine, such as drug discovery, and designing assignments that require higher-order thinking. At the same time, educators must discourage over-reliance on AI and encourage students to evaluate AI-generated responses critically \cite{abd2023large}. Students, in turn, should be aware of the limitations of LLMs, including privacy, copyright, transparency, and bias issues. They should use these tools in a manner that is both ethical and constructive, advancing their skills in medical practices while properly citing LLM use in their work and ensuring responsible usage of AI \cite{abd2023large, rudolph2023chatgpt}.\\
\indent Nevertheless, while emerging chatbots are expected to provide more accurate and safer content with genuine citations and fewer errors, the sufficiency of explainability and transparency under current and proposed international regulatory frameworks remains unclear \cite{EU2017, USFDA}. Therefore, one strategy to reduce inappropriate medical advice is to limit LLM training to controlled and validated medical texts. For instance, the GatorTronGPT, trained on 82 billion words of anonymized clinical text, demonstrates higher accuracy in answering medical questions than previous models \cite{yang2022large}. In addition, as medical research and documentation might not always be up-to-date or accurate, developers aiming for medical applications of their LLMs should implement quality management systems from the start. This aligns with the protocols defined in current regulatory frameworks and the future requirements of AI safety.\\
\indent To summarize, this section has outlined the ethical considerations and practices for the responsible use of LLM-based chatbots. We have emphasized the importance of safeguarding privacy and data, the necessity of ensuring language diversity and equal rights, and the establishment of academic and medical protocols. These measures are crucial for the ethical advancement of chatbot technologies, ensuring they are both equitable and beneficial across diverse user groups.
\vspace{-2mm}
\section{Conclusion}
\label{sec:Conclusion}
In this comprehensive survey, we have explored the realm of LLM-based chatbots. We begin with the formative years of chatbot evolution, then explore the LLMs, including their underlying architecture, working principles, and groundbreaking features, and later provide an overview of the existing and emerging LLM-based chatbots. Next, we examine the diverse applications across various fields, including education, research, healthcare, and others. Alongside their potential, we have also discussed the challenges they confront from technical, ethical, and misuse perspectives. Finally, we round off our survey by looking into the technical upgrades and ethical standards, highlighting their advancement towards greater efficiency, sustainability, and commitment to responsibility. As our survey concludes, we hope it serves as a valuable resource in the ever-evolving AI landscape, fostering discussions and reflections on the path to AGI and the role of LLMs therein.
% \section*{Acknowledgments}
% This should be a simple paragraph before the References to thank those individuals and institutions who have supported your work on this article.
\begin{comment}
    {\appendix[Proof of the Zonklar Equations]
Use $\backslash${\tt{appendix}} if you have a single appendix:
Do not use $\backslash${\tt{section}} anymore after $\backslash${\tt{appendix}}, only $\backslash${\tt{section*}}.
If you have multiple appendixes use $\backslash${\tt{appendices}} then use $\backslash${\tt{section}} to start each appendix.
You must declare a $\backslash${\tt{section}} before using any $\backslash${\tt{subsection}} or using $\backslash${\tt{label}} ($\backslash${\tt{appendices}} by itself
 starts a section numbered zero.)}
\end{comment}
% \section{References Section}
\bibliographystyle{IEEEtran}
\bibliography{references}
\begin{comment}
    \section{Simple References}
You can manually copy in the resultant .bbl file and set second argument of $\backslash${\tt{begin}} to the number of references
 (used to reserve space for the reference number labels box).
\end{comment}
%
\begin{comment}
    
\end{comment}
%
\begin{comment}
    \newpage

If you have an EPS/PDF photo (graphicx package needed), extra braces are
 needed around the contents of the optional argument to biography to prevent
 the LaTeX parser from getting confused when it sees the complicated
 $\backslash${\tt{includegraphics}} command within an optional argument. (You can create
 your own custom macro containing the $\backslash${\tt{includegraphics}} command to make things
 simpler here.)

\vspace{11pt}

\bf{If you will not include a photo:}\vspace{-33pt}
\begin{IEEEbiographynophoto}{John Doe}
Use $\backslash${\tt{begin\{IEEEbiographynophoto\}}} and the author name as the argument followed by the biography text.
\end{IEEEbiographynophoto}    
\end{comment}

\begin{IEEEbiography}[{\includegraphics[width=1in,height=1.25in,clip,keepaspectratio]{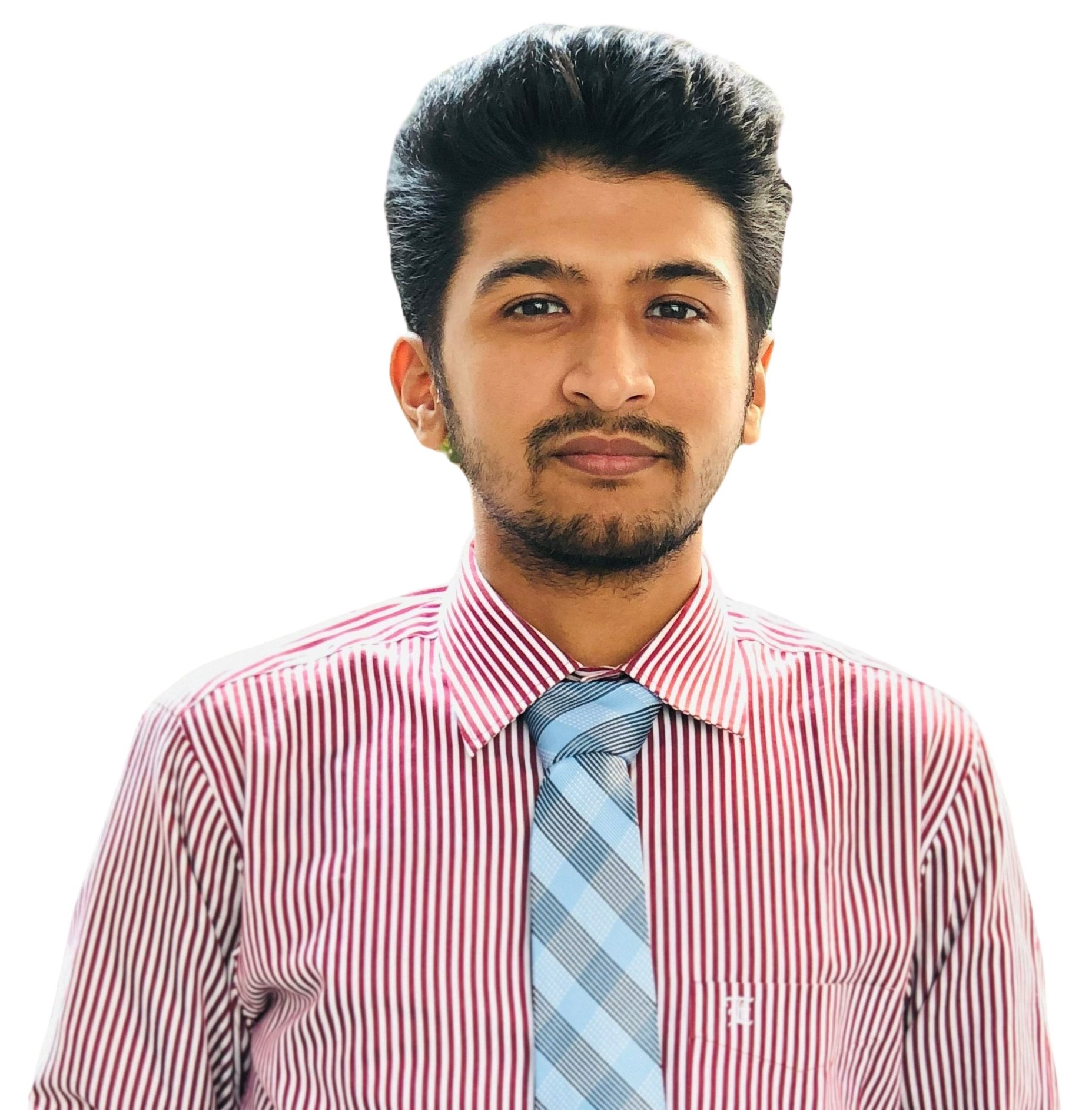}}]{Sumit Kumar Dam} is a Ph.D. Researcher at the Department of Artificial Intelligence at Kyung Hee University, South Korea. He received his B.Sc. Engineering degree in Computer Science and Engineering from Khulna University, Bangladesh, in 2020. His research interests include adversarial robustness, computer vision, self-supervised learning, and machine learning.
\end{IEEEbiography}
\begin{IEEEbiography}[{\includegraphics[width=1in,height=1.25in,clip,keepaspectratio]{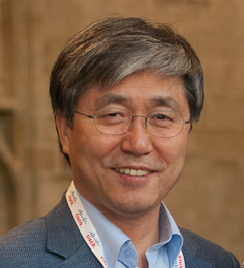}}] {Choong Seon Hong} (S’95-M’97-SM’11-F’23) received the B.S. and M.S. degrees in electronic engineering from Kyung Hee University, Seoul, South Korea, in 1983 and 1985, respectively, and the Ph.D. degree from Keio University, Tokyo, Japan, in 1997. In 1988, he joined KT in Gyeonggi-do, South Korea, where he was involved in broadband networks as a member of the Technical Staff. Since 1993, he has been with Keio University. He was with the Telecommunications Network Laboratory, KT, as a Senior Member of Technical Staff and as the Director of the Networking Research Team until 1999. Since 1999, he has been a Professor at the Department of Computer Science and Engineering at Kyung Hee University. His research interests include the future Internet, intelligent edge computing, network management, and network security. Dr. Hong is a member of the Association for Computing Machinery (ACM), the Institute of Electronics, Information and Communication Engineers (IEICE), the Information Processing Society of Japan (IPSJ), the Korean Institute of Information Scientists and Engineers (KIISE), the Korean Institute of Communications and Information Sciences (KICS), the Korean Information Processing Society (KIPS), and the Open Standards and ICT Association (OSIA). He has served as the General Chair, the TPC Chair/Member, or an Organizing Committee Member of international conferences, such as the Network Operations and Management Symposium (NOMS), International Symposium on Integrated Network Management (IM), Asia-Pacific Network Operations and Management Symposium (APNOMS), End-to-End Monitoring Techniques and Services (E2EMON), IEEE Consumer Communications and Networking Conference (CCNC), Assurance in Distributed Systems and Networks (ADSN), International Conference on Parallel Processing (ICPP), Data Integration and Mining (DIM), World Conference on Information Security Applications (WISA), Broadband Convergence Network (BcN), Telecommunication Information Networking Architecture (TINA), International Symposium on Applications and the Internet (SAINT), and International Conference on Information Networking (ICOIN). He was an Associate Editor of the IEEE Transactions on Network and Service Management, the IEEE Journal of Communications and Networks, and the International Journal of Network Management. Additionally, he was an Associate Technical Editor of the IEEE Communications Magazine. He currently serves as an Associate Editor of the International Journal of Network Management and the Future Internet Journal.
\end{IEEEbiography}
%\vspace{-18mm}
\begin{IEEEbiography}
[{\includegraphics[width=1in,height=1.25in,clip,keepaspectratio]{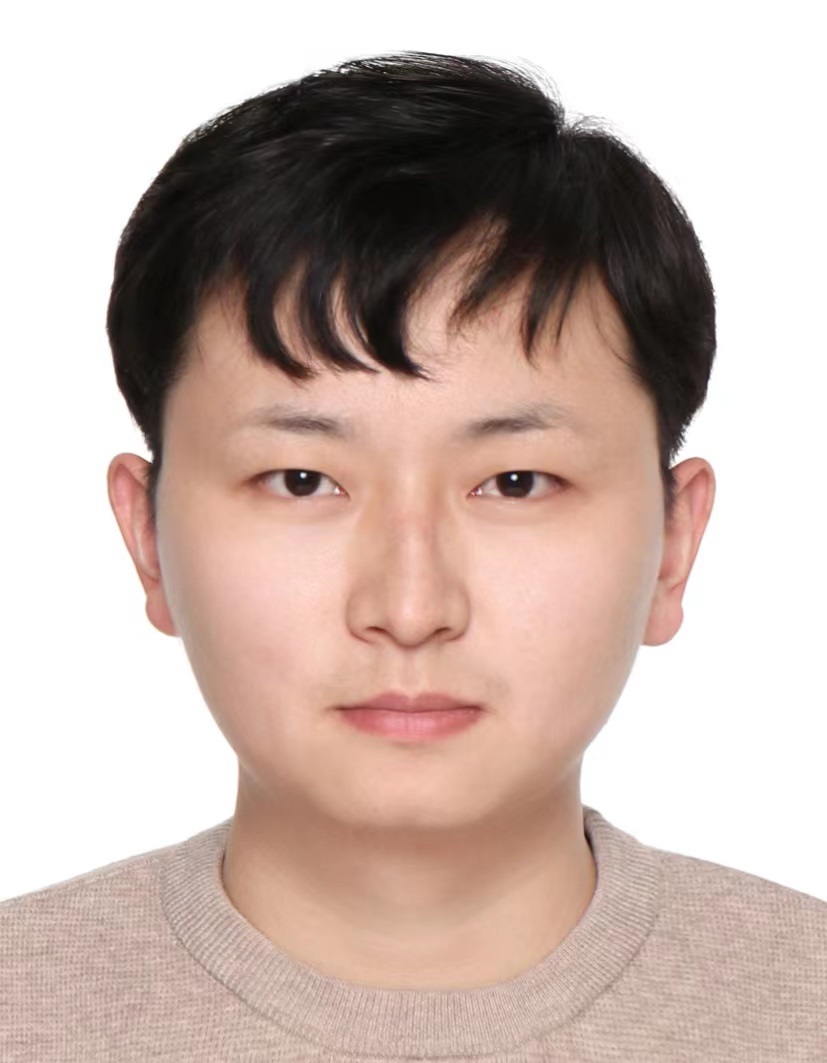}}]{Yu Qiao} (S’24) received his B.E. degree in Internet of Things Engineering and M.E. degree in Computer Science and Technology from Nanjing University of Information Science and Technology (NUIST), Nanjing, China, in 2016 and 2019, respectively. He is currently pursuing a Ph.D. degree at the Department of Artificial Intelligence at Kyung Hee University (KHU), South Korea. Before his Ph.D., he served as a Camera Software Engineer at Spreadtrum Communications (UNISOC), Shanghai, China, from 2019 to 2022. His interests include machine learning, federated learning, adversarial machine learning, self-supervised learning, and distributed edge intelligence.
\end{IEEEbiography}
%\vspace{-18mm}
\begin{IEEEbiography}[{\includegraphics[width=1in,height=1.25in,clip,keepaspectratio]{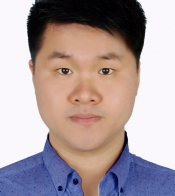}}]{Chaoning Zhang} received the B.E. and M.E. degrees in Electrical Engineering from Harbin Institute of Technology, China, in 2012 and 2015, respectively, and the Ph.D. degree from KAIST in 2021. Since 2022, he has been an Assistant Professor at the Department of Artificial Intelligence, School of Computing, Kyung Hee University. Before that, he worked as a post-doctoral researcher at KAIST. His research interests include but are not limited to adversarial machine learning and self-supervised learning for addressing model robustness and data efficiency issues in computer vision and beyond. 
\end{IEEEbiography}
\end{document}